\documentclass[lettersize,journal]{IEEEtran}
\usepackage{amsmath,amsfonts}
\usepackage{array}
\usepackage[caption=false,font=normalsize,labelfont=sf,textfont=sf]{subfig}
\usepackage{textcomp}
\usepackage{stfloats}
\usepackage{url}
\usepackage{verbatim}
\usepackage{graphicx}
\usepackage{times}
\usepackage{epsfig}
\usepackage{amssymb}
\usepackage{multirow}
\usepackage{booktabs}
\usepackage{setspace}
\usepackage{color}
\usepackage{algorithm}
\usepackage{algpseudocode}
\usepackage{algorithmicx}
\usepackage{hyperref}
\usepackage[table,xcdraw]{xcolor}
\usepackage{threeparttable}
\def\BibTeX{{\rm B\kern-.05em{\sc i\kern-.025em b}\kern-.08em
    T\kern-.1667em\lower.7ex\hbox{E}\kern-.125emX}}
\usepackage{balance}
\begin{document}
\title{Taming Self-Supervised Learning for Presentation Attack Detection: De-Folding and De-Mixing}

\author{Zhe Kong, Wentian Zhang, Feng Liu, Wenhan Luo, Haozhe Liu, Linlin Shen and Raghavendra Ramachandra
\thanks{This work was supported in part by the National Natural Science Foundation of China under Grant 62076163 and Grant 82261138629, in part by the Guangdong Basic and Applied Basic Research Foundation 2023A1515010688, in part by the Shenzhen Fundamental Research Fund under Grant JCYJ20190808163401646 and JCYJ20220531101412030, in part by Shenzhen Science and Technology Program under Grant JSGG20220831093004008, and in part by the Research Council of Norway through the Project “SALT” under Grant 332266.. (Zhe Kong and Wentian
Zhang contributed equally to this work.) (Corresponding author: Feng Liu.)}
\thanks{Zhe Kong, Wentian Zhang, Feng Liu, Haozhe Liu, and Linlin Shen are with the College of Computer Science and Software Engineering, Shenzhen University, Shenzhen 518060, China; SZU Branch, Shenzhen Institute of Artificial Intelligence and Robotics for Society, China, and also with Guangdong Key Laboratory of Intelligent Information Processing, Shenzhen University, Shenzhen 518060, China (email: liuhaozhe2019@email.szu.edu.cn; kongzhe2020@email.szu.edu.cn; feng.liu@szu.edu.cn; llshen@szu.edu.cn).}
\thanks{Wenhan Luo is with the School of Cyber Science and Technology, Shenzhen Campus of Sun Yat-sen University, Shenzhen, Guangdong 518107, P.R. China (email: whluo.china@gmail.com).}
\thanks{Raghavendra Ramachandra is with the Norwegian Biometrics Laboratory (NBL), Norwegian University of Science and Technology, 2818 Gjøvik, Norway (email: raghavendra.ramachandra@ntnu.no).}
\thanks{This work was done when Haozhe Liu was a visiting student at NTNU, Gjøvik, Norway}}

\markboth{IEEE TRANSACTIONS ON NEURAL NETWORKS AND LEARNING SYSTEMS}%
{}

\maketitle

\begin{abstract}
Biometric systems are vulnerable to Presentation Attacks (PA) performed using various Presentation Attack Instruments (PAIs). Even though there are numerous Presentation Attack Detection (PAD) techniques based on both deep learning and hand-crafted features, the generalization of PAD for unknown PAI is still a challenging problem. In this work, we empirically prove that the initialization of the PAD model is a crucial factor for the generalization, which is rarely discussed in the community. Based on such observation, we proposed a self-supervised learning-based method, denoted as DF-DM. Specifically, DF-DM is based on a global-local view coupled with \textbf{D}e-\textbf{F}olding and \textbf{D}e-\textbf{M}ixing to derive the task-specific representation for PAD. During De-Folding, the proposed technique will learn region-specific features to represent samples in a local pattern by explicitly minimizing generative loss. While De-Mixing drives detectors to obtain the instance-specific features with global information for more comprehensive representation by minimizing interpolation-based consistency. Extensive experimental results show that the proposed method can achieve significant improvements in terms of both face and fingerprint PAD in more complicated and hybrid datasets when compared with state-of-the-art methods. When training in CASIA-FASD and Idiap Replay-Attack, the proposed method can achieve an 18.60\% Equal Error Rate (EER) in  OULU-NPU and MSU-MFSD, exceeding baseline performance by 9.54\%. The source code of the proposed technique is available at \textsl{\href{https://github.com/kongzhecn/dfdm}{https://github.com/kongzhecn/dfdm}}.
\end{abstract}


\section{Introduction}

With the applications in mobile phone unlocking, access control, payment tool, and other security scenarios, biometric systems are widely used in our daily lives. Among the most popular biometric modalities, fingerprint and face play vital roles in numerous access control applications. However, several reported studies \cite{ramachandra2017presentation,singh2020survey} have demonstrated that the existing systems are easily spoofed by presentation attacks (PAs) made from low-cost materials and instruments, e.g. Rigid Mask for face \cite{heusch2020deep} and silica-gel for fingerprint \cite{9457215}. These issues raise wide concerns about the vulnerability of biometric systems incorporated in access control applications. Therefore, it is essential to detect presentation attacks to achieve reliable biometric applications.

To reliably address the vulnerability of the biometric systems to PAIs, several Presentation Attack Detection (PAD) methods have been proposed \cite{ramachandra2017presentation}, which can be divided into hardware and software-based methods.
Hardware-based solutions \cite{heusch2020deep,9335499,7018027} employ special types of sensors to capture liveness characteristics. For instance, Heusch et al. \cite{heusch2020deep} adopt short wave infrared (SWIR) imaging technology to detect face PAs, which shows superior performance over similar models working on color images. A light field camera (LFC) is introduced by Raghavendra et al.\cite{7018027} to detect PAs by exploring the variation of the focus between multiple depth images. For fingerprints, an optical coherence tomography (OCT)-based PAD system is designed by Liu et al. \cite{9335499} to obtain the depth information of fingerprints. Generally speaking, hardware-based solutions are sensor-specific, resulting in strong security but weak applicability because of usability or cost limitations, and the current mainstream is software-based methods.

\begin{figure*}
  \centering
  \includegraphics[width=.88\textwidth]{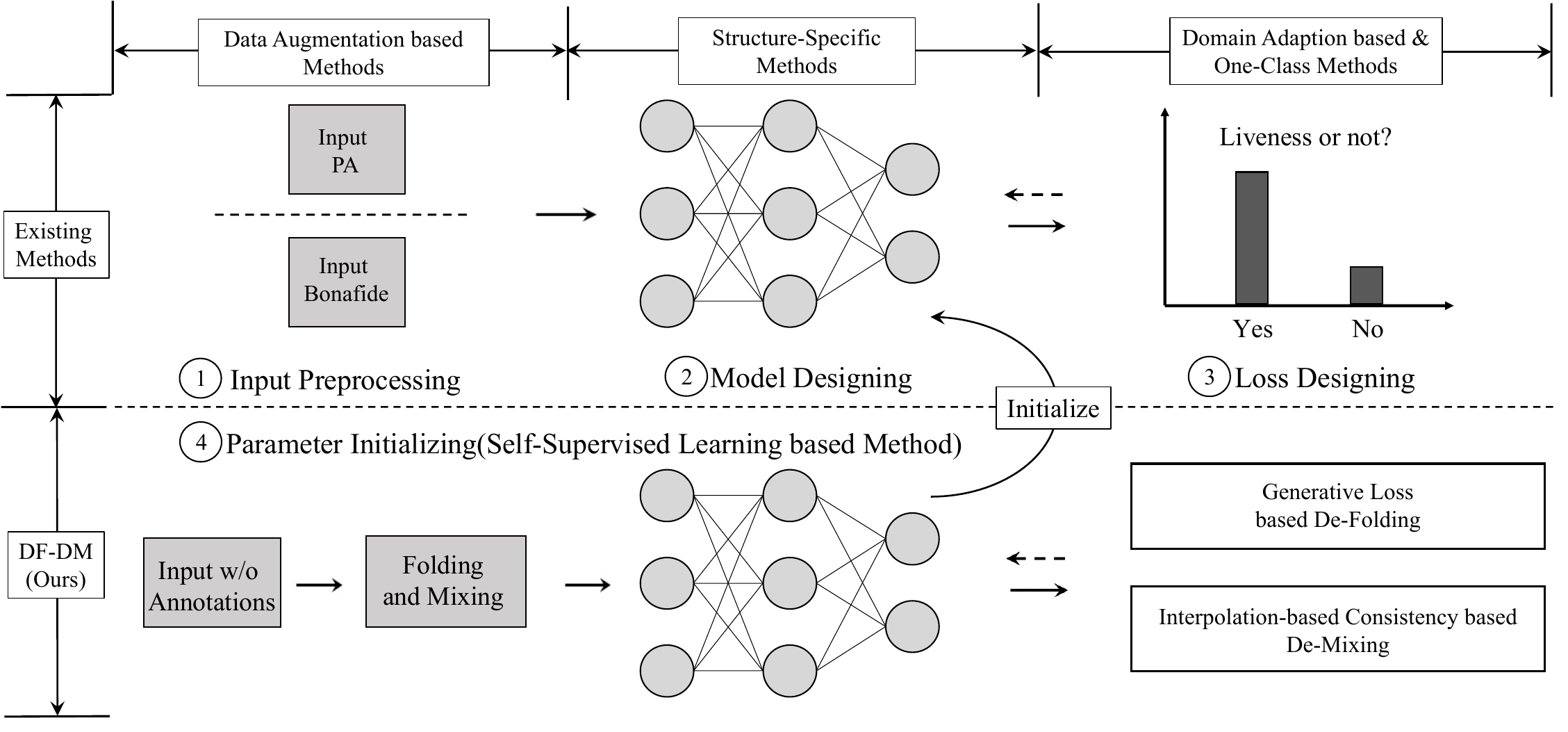}
  \caption{ The groups of software-based presentation attack detection. Group 1 is Input Preprocessing, Group 2 refers to Model Designing, and Group 3 is Loss Function Designing. Different from existing groups, the proposed method investigates the initialization of the PA detector, which can be concluded as an independent solution, i.e. Group 4: Parameter Initializing.
  }
  \label{fig:motivation}
\end{figure*}

\begin{table}[!htbp]
\caption{Average performance of PA Detector with Different Initialization in terms of EER (\%) $\downarrow$, AUC (\%) $\uparrow$ and TDR(\%)@FDR=1.0\% $\uparrow$ under the Cross-Dataset Setting of the face. More details can be found in the appendix.}
\centering
\label{tab:motivation}
\setlength\tabcolsep{6pt}
\resizebox{.49\textwidth}{!}{%
\begin{tabular}{cccc}
\hline
\multirow{2}{*}{Group1:}                       & \multicolumn{1}{c|}{OULU-NPU{[}7{]}} & \multirow{2}{*}{Group2:} & Idiap Replay-Attack{[}9{]} \\
                                               & \multicolumn{1}{c|}{MSU-MFSD{[}8{]}} &                          & CASIA-FASD{[}10{]}         \\ \hline
\multicolumn{1}{c|}{Metrics}                   & EER (\%)                             & AUC (\%)                 & TDR (\%)                   \\ \hline
\multicolumn{1}{c|}{Trained from Scratch}      & 42.3                                 & 59.51                    & 4.14                       \\ \hline
\multicolumn{1}{c|}{Pre-trained from ImageNet} & 31.28                                & 73.68                    & 10.06                      \\ \hline\rowcolor[HTML]{EFEFEF}
\multicolumn{1}{c|}{DF-DM}                     & \textbf{18.78}                       & \textbf{89.62}           & \textbf{30.39}             \\ \hline
\end{tabular}%
}
\end{table}

Fig. \ref{fig:motivation} illustrates the recent progress on the software-based PAD algorithms that can be categorized into three groups: 1) Input Preprocessing, 2) Model Design, and 3) Loss Function. In the case of Input Preprocessing \cite{liu2018learning,8616677,9079541,wang2021rgb,guo2019improving}, Larbi et al. \cite{8616677} propose a model, namely DeepColorFASD model, which adopts various color spaces (RGB, HSV, YCbCr) as input to achieve the reliable performance of PAD. Despite the improvement, the additional color spaces need to be processed, leading to extra computation.
Unlike Input Preprocessing, the  Model Designing approaches pay more attention to the specific CNN-based architectures \cite{zhang2022multi}. Many prior works adopt hand-crafted features such as LBP \cite{de2013can}, HoG \cite{yang2013face}, SIFT \cite{patel2016secure}, and Surf \cite{boulkenafet2016face}, then employ traditional classifiers such as LDA and SVM. But hand-crafted features are sensitive to noise  and illumination, resulting in poor generalization performance. Consequently, structure-specific methods based on convolution neural networks (CNN) are proposed. In particular, Liu et al. \cite{liu2018learning} propose a CNN-RNN model to learn auxiliary features, including depth and rPPG, for PAD.
To tackle the cross-domain issue, current works are trying to improve the generalization by training models with specific learning objectives \cite{wang2019domain}. Jia et al. \cite{jia2020single} propose an Asymmetric Triplet Loss to mine the PAD features and design a Single-Side Adversarial Loss to align the mined features between the different domains.

For these software-based PAD algorithms, an important interference factor is the initialization of the PA detector. Generally, training from scratch and pre-training using ImageNet \cite{russakovsky2015imagenet} are two common methods. In terms of PAD, it is challenging to collect large-scale data; hence, without any prior knowledge, it is difficult to train the model from scratch (i.e., random initialization) to learn discriminative features. As listed in Table \ref{tab:motivation}, without any pre-training strategies, the detector can only reach 42.30 \% Equal Error Rate (EER) for face anti-spoofing. While the detector pre-trained from ImageNet can achieve 31.28 \% EER, performing higher generalization against different datasets. Such empirical results indicate that initialization plays a vital role in improving the generalization of PAD. However, taking a pre-trained model from ImageNet as initialization is also not a proper choice. As a large-scale dataset, the cost of time and computation carried on ImageNet is an over-heavy workload to train new proposed PAD CNN architectures. Meanwhile, face and fingerprint images are quite different from natural images in both texture and context, and the pre-trained model is thus not an ideal and reasonable starting point for the PAD task.
\begin{figure*}
  \centering
  \includegraphics[width=.96\textwidth]{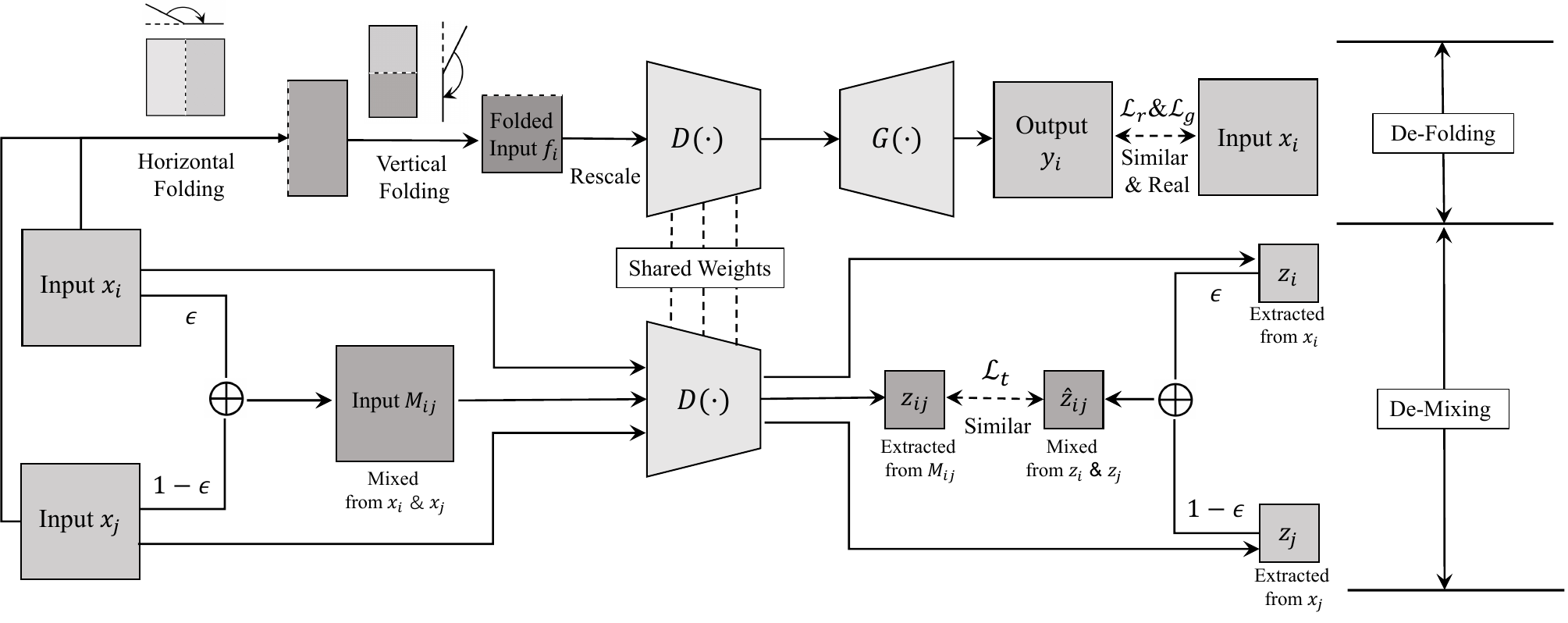}
  \caption{The pipeline of the proposed self-supervised learning-based method (DF-DM). The feature extractor $D(\cdot)$ is trained by De-Folding and De-Mixing tasks simultaneously. In the De-Folding task, the input image is folded, and the model's target is to reconstruct the image by minimizing the generative losses. In the De-Mixing task, two images are mixed by a random weight $\epsilon$, and the learning objective is to guarantee the Interpolation-based Consistency between the input space and feature space embedded by $D(\cdot)$. Note that both $x_i$ and $x_j$ are employed for De-Folding, but only the pipeline of $x_i$ is shown in the figure due to the same processing for $x_i$ and $x_j$.
  }
  \label{fig:pipeline}
\end{figure*}

In order to solve the addressed problems, a self-supervised learning method, denoted as DF-DM, is proposed in this paper. 
Without any PAD labels, two pretext tasks are designed to train the network for the initialization of the PA detector. 
Based on the chirality of fingerprints and the symmetry of faces, a generative task denoted as De-Folding,
is designed to force a CNN-based model to reconstruct the folded images by learning the specific patterns among various regions in faces or fingerprints. 
To facilitate the extraction of global features, another pretext task De-Mixing is proposed.

In the De-Folding task, the texture of fingerprints or faces is all overlapped in the folded images. The network is trained to dissociate the chaotic texture in the folded images. In this process, the network learns texture-related semantics, which is helpful for presentation attack detection tasks. Due to the chirality of fingerprint images, a fingerprint image is folded in the horizontal and vertical directions, respectively. But the face is left-right symmetrical, and the face images are only folded once in the vertical direction to learn the symmetrical features of the face. In the De-Folding task, the model learns to represent the images with region-specific features in a local view.
To further strengthen the representation between samples, the De-Mixing task is proposed to facilitate the extraction of global features. In the De-Mixing task, the network disentangles the mixed images from two samples, which learns the texture-related semantic information in a global view. The pretext task De-Folding and De-Mixing reconstruct images from local and global views respectively. The detector can achieve 18.78\% EER after using our proposed DF-DM pretext task, which gets a much better result than the model trained from scratch and pre-trained from ImageNet. The proposed method investigates the initialization of the PA detector, and it can be categorized into an independent solution in Fig.\ref{fig:motivation}: Parameter Initializing. It should be noted that the proposed parameter initializing strategy could be used for other network architecture. Without any extra training samples and PAD labels, the proposed methods can obtain an ideal initialization, which can improve fingerprint PAD baseline from 73.92\% to 90.96 \% in TDR (True Detection Rate)@ FDR (False Detection Rate)=1\% and promote the EER (Equal Error Rate) of face PAD baseline by 8.12 \%. In summary,

\begin{itemize}
\item A generative task, De-Folding, is designed for fingerprints and faces PAD to explore the specific patterns and textures among different regions.
\item As a complementary task, the De-Mixing task with interpolation-based loss is proposed to learn more global features to better represent images.
\item The proposed method, which performs in an unsupervised manner with limited computational resources, achieves impressive performance in terms of face and fingerprint PAD.
\end{itemize}

\section{Related Works}

Since this paper mainly focuses on PAD solutions based on self-supervised learning. In this section, we review not only the most representative works of self-supervised learning in vision tasks but also other fingerprint and face PAD methods.

\subsection{Fingerprint Presentation Attack Detection Methods}

For fingerprint presentation attack detection methods, CNN-based networks can achieve satisfactory performance \cite{nogueira2016fingerprint,8306930,pereira2020robust,9079541,9457215,liu2022fingerprint}. Nogueira et al. \cite{nogueira2016fingerprint} pre-train deep CNNs for object recognition and then fine-tune the CNN for fingerprint PAD. Chugh et al. \cite{8306930} extract local patches centered and align the input using minutiae for the fingerprint PAD Model. To improve the generalization, researchers have further proposed numerous methods to improve the performance across ``unknown” or novel attacks. Pereira et al. \cite{pereira2020robust} address the generalization problem by applying a regularisation technique based on adversarial training. A Generative Adversarial Network (GAN) based data augmentation, called Universal Material Generator (UMG), is proposed by Chugh et al. \cite{9079541} to transfer the style (texture) characteristics between fingerprint images to train a robust PA detector. Liu et al. \cite{9457215} propose a global-local model-based PAD (RTK-PAD) method to overcome information loss and improve generalization ability.

One-class based approach is also proposed to address unknown attack problems. In \cite{sequeira2015fingerprint}, the authors compare the performance of supervised and semi-supervised approaches that rely solely on bona fide samples. Liu et al. \cite{9335499} propose a one-class PAD model OCPAD, which is based on an autoencoder network. The proposed OCPAD model is learned from the training set containing only bona fides samples. The reconstruction error and latent code obtained by the trained autoencoder network are used to calculate the spoof score.

\subsection{Face Presentation Attack Detection Methods}

Software-based face presentation attack detection methods can be categorized into hand-craft-based methods and deep learning-based methods. LBP \cite{de2013can}, HoG \cite{yang2013face}, SIFT \cite{patel2016secure}, and Surf \cite{boulkenafet2016face} are often used to extract hand-craft features, and then these extracted features are input to a traditional classifier such as LDA and SVM for classification. Hand-crafted features are sensitive to noise, so they cannot generalize well to different illumination or different attack types.

With the success of CNN, numerous deep learning-based face PAD methods have been proposed \cite{yu2020face,qin2020learning,wang2020deep,yu2020fas,yu2021revisiting,qin2021meta,yu2021transrppg,wang2022domain,yu2021dual,yu2022benchmarking,cui2022rethinking,liu2018learning,liu2019deep,george2020learning,jia2020single,yu2020searching,liu2021adaptive,yang2019face}. Liu et al. \cite{liu2018learning} propose a CNN-RNN model to learn auxiliary features, including depth and rPPG, for PAD. Yang et al. \cite{yang2019face} proposed a data collection method using a data synthesis technique to generate spoof samples. A Deep Tree Network (DTN)  is proposed by Liu et al. \cite{liu2019deep} to partition the spoof samples into semantic sub-groups and detect PAs by routing test samples into similar clusters. Yu et al. \cite{yu2020searching} propose a Central Difference Convolution (CDC) layer to capture intrinsic detailed patterns via aggregating both intensity and gradient information. By adopting Neural Architecture Search (NAS), the CDC-based network can achieve superior performance. The one-class loss proposed by George et al. \cite{george2020learning} tries to learn a compact embedding space for the bona fide samples. However, the one-class loss only considers the domain-invariant features and ignores the differences among domains. Hence, Jia et al. \cite{jia2020single} propose an Asymmetric Triplet Loss to mine the PAD features and design a Single-Side Adversarial Loss to align the mined features between the different domains. Besides, Wang et al. \cite{wang2022domain} proposed a novel Shuffled Style Assembly Network (SSAN) to extract and reassemble different content and style features for a stylized feature space. More recently, Zhang et al. \cite{zhang2022effective} propose to extract the prior knowledge from the face-related works in a face system  to improve the generalization of face PAD.

\subsection{Self-supervised Learning Methods for Vision Tasks}

Self-supervised learning refers to learning methods in which CNNs are trained with automatically generated labels and then transferred to other computer vision tasks \cite{jing2020self,zhao2020review,liu2022self,ma2020self}. Based on the categories of the generated labels, self-supervised learning can be roughly divided into generative and contrastive learning \cite{liu2021anomaly}. However, both kinds of methods cannot be directly used in PAD. As fingerprint and face images for recognition lack colorful information, and generally with low resolution,  generative learning, such as image colorization \cite{zhang2016colorful} and image super-resolution \cite{ledig2017photo}, cannot be conducted in this case.
Meanwhile, for image in-painting \cite{pathak2016context}, and GANs \cite{goodfellow2014generative,zhu2017unpaired}, large-scale data is required to establish a compact feature space, while the dataset for PAD cannot meet the requirement.
On the other side, face images have strong spatial specifications after alignment, while the spatial relation of fingerprints is weak.
Hence, contrastive learning, like predicting the relative position \cite{doersch2015unsupervised} and rotation \cite{gidaris2018unsupervised}, is easy for the face but too hard for fingerprint.
Another group of contrastive learning is instance discrimination, like MoCo \cite{he2020momentum,chen2020improved}, SimCLR \cite{chen2020simple} and BYOL \cite{grill2020bootstrap}. Through embedding each instance/image into different classes, the mentioned studies have shown solid improvement in natural images. However, in the PAD task, the bijective relation between each image and prediction leads the model to learn identification rather than PAD features, resulting in poor generalization performance \cite{wang2020cross}.

Self-supervised learning methods for PAD have been proposed \cite{wang2022patchnet,muhammad2022self,wang2021consistency}. Wang et al. \cite{wang2022patchnet} reformulate face anti-spoofing as a fine-grained patch-type recognition task and present a simple training framework called PatchNet to efficiently learn the embedding of the patch with the spoof-related capture characteristics. Wang et al. \cite{wang2022patchnet}, and our proposed method all emphasize that texture or structural materials play an important role in PAD. But PatchNet adopts a recognition task to learn these texture or materials features, while we propose a reconstruction task (De-Folding) to learn the texture and material features.
Furthermore, Wang et al. \cite{wang2021consistency} propose a novel embedding-level and prediction-level consistency regularization method for deep face anti-spoofing. The consistency of the two feature maps, extracted from the same input but with different augmentation, is then employed to boost the PAD model. But in our De-Mixing tasks, we compute the distance of two feature maps to learn the relationship between the samples, and the two feature maps in the De-Mixing task are derived from two different input images. When the training data are videos, Muhammad et al. \cite{muhammad2022self} propose Temporal Sequence Sampling (TSS) for 2D face PAD by removing the estimated inter-frame 2D affine motion in the view and encoding the appearance and dynamics of the resulting smoothed video sequence into a single RGB image. 

In this paper, a novel self-supervised learning, namely DF-DM, is proposed to improve the performance of PAD. Unlike existing PAD methods, the proposed method is free of any PAD labels and extra data and pays more attention to the specification of the face and fingerprint. Two pretext tasks,  De-Mixing and De-Folding, are proposed to search for a reasonable initialization for the PA detector. The De-Folding task explores the properties of the face and fingerprint, such as chirality and symmetry, by searching the differences among the patches from a given sample. In contrast, De-Mixing requires the model to embed the samples into a compact but distinguishable feature space by localizing the relationship between the different samples. By drawing De-Folding and De-Mixing simultaneously, adequate PAD features are extracted, which can be useful for detecting PAs. Extensive experiments clearly show significant improvements in the performance of face and fingerprint PAD.

\section{Proposed Method}
\label{sec:method}

Figure \ref{fig:pipeline} presents the block diagram of the proposed DF-DM, which adopts De-Folding and De-Mixing to reliably capture the  hierarchical  features useful for PAD. The goal of De-Folding is to reconstruct the raw image from the folded image.  Since the folded image and the corresponding ground truth can be easily obtained, the model in this task is directly trained by minimizing the generative losses in an explicit way. While the De-Mixing task is an ill-posed problem, where a single mixed image corresponds to two different images (irrespective of order). Hence, we introduce a new loss function called Interpolation-based Consistency to train the model for De-Mixing in an implicit way. In the following sections, we will present a detailed discussion of the proposed method.

\subsection{De-Folding Task: Searching Differences among the Patches}
\label{sec:ccdf}
The patterns of the face and fingerprint are quite different from that of natural images. A typical case of the point is that the fingerprint and face perform symmetric distribution in the global view of the images but chirality in the local patterns, such as texture features and reflection. In PAD, print photos, replay videos, and 3D masks are typical attacks for biometric recognition systems. Although the attacks are similar to the bona fide samples from the view of human vision, the texture of the attacks is generally unusual, with anomalous reflection due to the specification of the instruments.
As the features of PAD are mainly identical to the chiral features, a chirality-related pretext task, denoted as De-Folding, is proposed in this paper.

\begin{figure}[h]
  \centering
  \includegraphics[width=.48\textwidth]{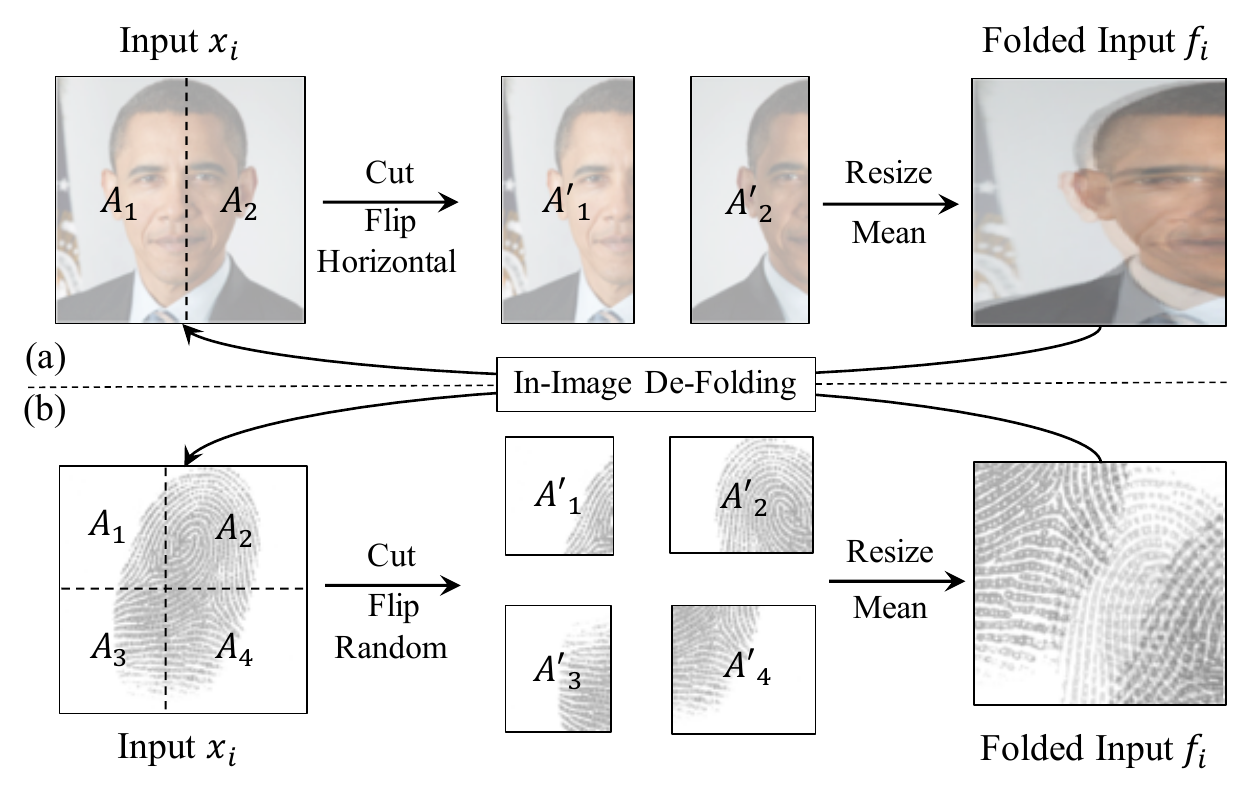}
  \caption{The pipeline of folding inputs in the De-Folding task. (a) presents the case of the face, and (b) refers to the fingerprint case. The fingerprint sample is selected from LivDet2017 \cite{mura2018livdet}, and the face is cropped from \cite{gecer2019ganfit}.
  }
  \label{fig:IF}
\end{figure}

As shown in Fig. \ref{fig:IF}, based on different modalities, we propose two strategies to fold images. In the case of the face, a vertical line is adopted to cut the input image $x_i$ into two patches, $A_1$ and $A_2$, which are then randomly selected to flip horizontally to obtain $A'_{1}$ and $A'_{2}$. Through resizing and averaging $A'_{1}$ and $A'_{2}$, the folded image $f_i$ can be calculated, which is then drawn as the input of the following part. Unlike the left-right symmetrical face, the fingerprint shows chirality in vertical and horizontal directions. Hence, $x_i$ is cropped into four patches, $\{A_1,A_2,A_3,A_4\}$, by the vertical and horizontal lines. And the flips with various directions are correspondingly adopted to generate $\{A'_1,A'_2,A'_3,A'_4\}$. In order to improve the difficulty of the task and prevent the model from overfitting, the lines for cutting are randomly localized rather than frozen in the middle of the image.

Since the paired data, i.e. ($f_i$,$x_i$) for De-Folding, can be generated easily, the model is trained explicitly by minimizing the generative losses. In particular, given $f_i$ as input, a feature extractor $D(\cdot)$ is adopted to embed $f_i$ into a latent representation $z_i$, while a generator $G(\cdot)$ is employed to reconstruct $z_i$ to $y_i$. By following such cycle pipeline $x_i \rightarrow f_i \rightarrow y_i$, $D(\cdot)$ and $G(\cdot)$ are trained end-to-end
by the learning objective,
\begin{equation}
  \label{eq:1}
\begin{aligned}
    &\min_{G,D} \underset{\substack{x_i \sim \mathcal{X}_t \\ f_i \sim  \mathcal{T}(\mathcal{X}_t)}}{\mathcal{L}_r}(y_i,x_i) +  \underset{\substack{x_i \sim \mathcal{X}_t \\ f_j \sim  \mathcal{T}(\mathcal{X}_t)}}{\mathcal{L}_g}(y_j,x_i) \\
    &\underset{\substack{x_i \sim \mathcal{X}_t \\ f_i \sim  \mathcal{T}(\mathcal{X}_t)}}{\mathcal{L}_r}(y_i,x_i) \!=\! \underset{\substack{x_i \sim \mathcal{X}_t \\ f_i \sim  \mathcal{T}(\mathcal{X}_t)}}{\mathbb{E}}\! || y_i - x_i ||_2 \\
    &\underset{\substack{x_i \sim \mathcal{X}_t \\ f_j \sim  \mathcal{T}(\mathcal{X}_t)}}{\mathcal{L}_g}(y_j,x_i) \!=\!
     \underset{x_i \sim \mathcal{X}_t}{\mathbb{E}}\![F(x_i)] - \underset{f_j \sim \mathcal{T}(\mathcal{X}_t)}{\mathbb{E}}\![F(y_j)]
\end{aligned}
\end{equation}
where $\mathcal{T}(\cdot)$ is the image folding and $\mathcal{X}_t$ refers to the training set.
$F(\cdot)$ is a discriminator, which has the same architecture with $D(\cdot)$ and is trained by maximizing $\mathcal{L}_g$. $\mathcal{L}_r$ leads $y_i$ to be similar with $x_i$ in a supervised mode, while $\mathcal{L}_g$ adopt the loss of WGAN \cite{arjovsky2017wasserstein} to ensure the realness of $y_i$ by following unsupervised setting.

\subsection{De-Mixing Task: Preserving the Relationship among Different Samples}
\label{sec:tcdm}
Since De-Folding performs in a local view, the model leans to represent the images with region-specific features for reconstruction.
Such a pretext task pays more attention to the local patterns but neglects the relationship between the samples. As a result, varying samples can be embedded into similar representations. However, PAD is a binary classification task in which the ideal embedding space is compact but distinguishable for different samples. Therefore, in this paper, another pretext task, denoted as De-Mixing, is proposed to further enhance the discrimination among the different samples. Inspired by the work \cite{liu2022robust,liu2021manifold},  the model is not only required to reconstruct folded images but also to disentangle mixed images from different samples.
In De-Mixing task, two samples $x_i$ and $x_j$ are mixed into $M_{ij}$ by
\begin{align}
  \label{eq:2}
M_{ij} = \underset{{\epsilon \sim \mathbb{U}(0,1)}}{\Delta}(x_i, x_j) =\epsilon  x_i + (1-\epsilon)  x_j
\end{align}
where $\mathbb{U}(0,1)$ is the uniform distribution from 0. to 1.,and $\epsilon$ is a scalar sampled from $\mathbb{U}(0,1)$ for mixing. Given $M_{ij}$ as input, the feature extractor $D(\cdot)$ is required to disentangle $M_{ij}$ into $x_i$ and $x_j$. However, such a requirement makes the task an ill-posed problem, which is hard to train end-to-end. Considering the groundtruth of De-Mixing, both $\{ x_i,x_j\}$ and $\{x_j,x_i\}$ are the correct results. But for $D(\cdot)$, the order changing in the ground truth is regarded as different labels. To overcome the problem, the De-Mixing task is trained in an implicit way using Interpolation-based Consistency $\mathcal{L}_t$,
\begin{equation}
    \label{eq:3}
\begin{aligned}
  \underset{x_i,x_j \sim \mathcal{X}_t}{\mathcal{L}_t}(x_i,x_j) &=
  \underset{x_i,x_j \sim \mathcal{X}_t}{\mathbb{E}}||z_{ij} - \hat{z}_{ij} + \delta||_2 \\
  \hat{z}_{ij} &= \Delta(z_i,z_j)
\end{aligned}
\end{equation}
where $\epsilon$ for $M_{ij}$ and $\hat{z}_{ij}$ is identical, $z_i$, $z_j$ and $z_{ij}$ are the outputs of $D(x_i)$, $D(x_j)$ and $D(M_{ij})$ respectively and $\delta$ is a random noise sampled from a Gaussian distribution with 0. mean and 0.1 standard deviations.  By minimizing the distance between $z_{ij}$  and $\hat{z}_{ij}$, the mixing operation is identical in both image and embedding space.
Since $\{z_i,z_j,z_{ij}\}$ has the same topological structure with $\{x_i,x_j,M_{ij}\}$, $M_{ij}$ can be de-mixed easily in the embedding space of $D(\cdot)$, which approximately meets the target of De-Mixing. Note that Interpolation-based Consistency has a trivial solution, e.g., embedding the same code for all images, $\delta \in \mathcal{N}(0,0.1)$ is thus added into $\mathcal{L}_t$ to enhance the gradients against collapsing cases.

\subsection{DF-DM based Presentation Attack Detection}
\label{sec:PAD}
Considering the complementarity between De-Folding and De-Mixing, the proposed method trains $D(\cdot)$ with both pretext tasks simultaneously, and the total learning objective can be concluded as
\begin{align}
        &\min_{G,D} \underset{\substack{x_i \sim \mathcal{X}_t \\ f_i \sim  \mathcal{T}(\mathcal{X}_t)}}{\mathcal{L}_r}\!\!\!\!\!\!(y_i,x_i) +\!\!\!\!\!\!  \underset{\substack{x_i \sim \mathcal{X}_t \\ f_j \sim  \mathcal{T}(\mathcal{X}_t)}}{\mathcal{L}_g}\!\!\!\!\!\!(y_j,x_i) +\!\!\!\!\!\! \underset{x_i,x_j \sim \mathcal{X}_t}{\mathcal{L}_t}\!\!\!\!\!\!(x_i,x_j)
\end{align}
After training, $D(\cdot)$ is employed as the initialization for a presentation attack detector $H(\cdot)$. Compared with $D(\cdot)$, $H(\cdot)$ has an additional fully-connected layer to map $z_i$ into a single scalar $v_i$, i.e., spoofness score. The spoofness score reflects the category probability (PA or not) of the given sample $x_i$. $H(\cdot)$ is trained through a common cross entropy-based objective as follows:
\begin{align}
  \label{eq:5}
  \underset{x_i \in \mathcal{X}_t}{\mathcal{L}_c}\!\!\!(x_i,v_i) \!\!=\!\! - \!\! \underset{x_i \in \mathcal{X}_t}{\mathbb{E}}[u_i log(v_i) + (1-u_i) log(1-v_i)]
\end{align}
where $v_i = H(x_i)$ and $u_i$ is the category annotation of $x_i$. For clarity, the proposed method is summarized in Algo. 1.

\begin{algorithm}
  \label{algo:1}
	\caption{Presentation Attack Detection using DF-DM}
	\begin{algorithmic}
    \Require\\
     Feature Extractor $D(\cdot)$; Generator $G(\cdot)$; Discriminator $F(\cdot)$; Training Set $\mathcal{X}_t$; Presentation Attack Detector $H(\cdot)$;
    \Ensure \State Trained $H(\cdot)$;
  \end{algorithmic}
  \begin{algorithmic}[1]
    \While{ $D(\cdot)$ has not converged}
         \For{$x_i$,$x_j$ in $\mathcal{X}_t$ }
            \State Derive folded input $f_i$ from $\mathcal{T}(x_i)$;
            \State Reconstruct $f_i$ to $y_i$ through $G(D(f_i))$;
            \State Update $G(\cdot)$ and $D(\cdot)$ by minimizing Eq.(\ref{eq:1});
            \State Update $F(\cdot)$ by maximizing $\mathcal{L}_g$;
            \State Calculate $M_{ij}$ from $x_i$ and $x_j$ through Eq.(\ref{eq:2});
            \State Update $D(\cdot)$ by minimizing Eq.(\ref{eq:3})
         \EndFor
 \EndWhile
 \While{ $H(\cdot)$ has not converged}
 \State Adopt $D(\cdot)$ as the initialization of $H(\cdot)$;
 \For{$x_i$ in $\mathcal{X}_t$ }
    \State Obtain spoofness score $v_i$ by $H(x_i)$;
    \State Update $H(\cdot)$ by minimizing Eq.(\ref{eq:5});
 \EndFor
  \EndWhile
\State Return $H(\cdot)$;
	\end{algorithmic}
\end{algorithm}

\section{Experimental Results and Analysis}
To evaluate the performance of the proposed method,
extensive experiments are carried on the publicly-available datasets, including LivDet2017\cite{mura2018livdet}, OULU-NPU \cite{boulkenafet2017oulu}, CASIA-FASD \cite{zhang2012face}, Idiap Replay-Attack \cite{chingovska2012effectiveness}, MSU-MFSD \cite{wen2015face}, Rose-Youtu \cite{li2018unsupervised,li2022one} and WMCA \cite{george2019biometric}. We first introduce the datasets and the corresponding implementation details. Then, the effectiveness of the proposed method is validated by analyzing the contribution of each component. Since this is the first time to adopt self-supervised learning for PAD, we finally compare the proposed method with both existing self-supervised methods and PA detectors to further prove the superiority of the proposed method.
\subsection{Datasets and Implementation Details}
As the proposed method is evaluated in two modalities, including fingerprint and face, we separately introduce the details of the corresponding protocols as follows:

\textbf{Fingerprint.} Due to the complete experimental settings, LivDet 2017 \cite{mura2018livdet} is used to test the methods on fingerprint PAD. Table A.1 in the appendix summarizes the information of LivDet2017, which is used to evaluate the performance of fingerprint PAD. The dataset comprises over 17,500 fingerprint images captured from three readers, i.e., Green Bit, Orcanthus, and Digital Persona. Specifically, GreenBit is employed for Italian border controls and insurance of Italian electronic documents. Orcanthus is widely used in the personal computer (PC). And Digital Persona is adopted in mobile devices, such as Nexus 7 tablet. Hence, the adopted readers are reliable for testing the practical performance of fingerprint PAD. For each sensor, about 1760 fingerprint images are used for training, 440 images for validation, and  3740 images for testing. To evaluate the generalization of the competing methods, cross-material and cross-sensor settings \cite{9457215} are used in this paper. For cross-material cases, the spoof materials available in the test set are deemed as unknown materials, which are inaccessible during training. The partition of materials follows the setting in \cite{9457215}. In the cross-sensor protocol, PA detectors are trained by the images collected using a randomly-selected sensor and then tested using the images from the other sensors. Equal Error Rate (EER), Area Under Curve (AUC), and true detection rate (TDR) @ false detection rate (FDR)=1\% are used to evaluate the performance of detection.

In terms of network architecture, MobileNet V2 \cite{sandler2018mobilenetv2} is selected as the backbone for the feature extractor and discriminator, while the corresponding generator is designed by following U-Net architecture \cite{ronneberger2015u}. 
Note that, in order to test the capacity of feature extraction and reduce the dependence on data scale, only the training set adopted for PAD is drawn to train the proposed method.

We compare the proposed method with both self-supervised learning-based methods and presentation attack detectors. For self-supervised learning-based methods, GAN-based discriminator \cite{isola2017image}, and auto-encoder based encoder \cite{isola2017image} is set as the baseline of the generative learning, while MoCo V2 \cite{chen2020improved} is selected as the representative method of the contrastive learning. Regarding PA detector, LivDet 2017 winner \cite{mura2018livdet}, and FSB \cite{8306930} are adopted as the competing method. For a more comprehensive analysis of the proposed method, multiple models-based PA detectors, including RTK-PAD \cite{9457215} and FSB + UMG Wrapper \cite{9079541}, are also included for reference.

\textbf{Face.} To test the performance of face PAD, 4 datasets (details have shown in the appendix), including OULU-NPU \cite{boulkenafet2017oulu} (denoted as O), CASIA-FASD \cite{zhang2012face} (denoted as C), Idiap Replay-Attack \cite{chingovska2012effectiveness} (denoted as I) and MSU-MFSD \cite{wen2015face} (denoted as M) are adopted in this paper for evaluation using two cross-dataset protocols. In Protocol-1, [O, M] and [C, I] are set as two groups, and the model is trained in one group and tested in the other. While in Protocol-2, three datasets are used to train, and the other dataset is adopted for evaluation. In these two protocols, we use the whole dataset in O, C, I, and M to train and test. For each video, only one randomly-selected frame is used to train or test the detectors. In particular, printed photos, display photos, and replayed videos are used to attack facial recognition systems. Various acquisition devices, such as laptops and smartphones, are considered in different datasets. Hence the robustness and generalization of the PAD methods can be tested through the cross-dataset protocol.

\begin{table*}[!htbp]
    \centering
\caption{Performance of Various PAD Methods with or without the Proposed Method as Initialization in terms of EER(\%) $\downarrow$, AUC(\%) $\uparrow$ and TDR(\%)@FDR=1.0\% $\uparrow$ under the Cross-Dataset Setting of Face  (Protocol-1).  }
\label{tab:face_compare}
\setlength\tabcolsep{6pt}
\resizebox{.98\textwidth}{!}{
\begin{threeparttable}
\begin{tabular}{c|cccccc|ccc}
\hline

                        & \multicolumn{3}{c|}{{[}O,M{]} to {[}C, I{]}}                                                                       & \multicolumn{3}{c|}{ {[}C,I{]} to {[}O, M{]}}                           & \multicolumn{3}{c}{\textbf{ Mean $\pm$ s.d.}}                                                              \\ \cline{2-10}

                        &  EER(\%)            &  AUC(\%)            & \multicolumn{1}{c|}{ TDR(\%)} &  EER(\%)            &  AUC(\%)            &  TDR(\%)  &  EER(\%)                   &  AUC(\%)                   &  TDR(\%)          \\ \hline

DeepPixBiS \cite{george2019deep}    &  22.93          &  79.13          &  0.00                                                      &  22.45          &  85.70          &  24.37          &  22.69 $\pm$ 0.34          &  82.42 $\pm$ 4.65          &  12.19 $\pm$ 17.23          \\
\rowcolor[HTML]{EFEFEF}
 Ours:DeepPixBiS + DF-DM &  \textbf{15.94} &  \textbf{92.60} &  \textbf{42.34}                                             &  \textbf{22.14} &  \textbf{86.36} &  \textbf{34.08} &  \textbf{19.04 $\pm$ 4.38} &  \textbf{89.48 $\pm$ 4.41} &  \textbf{38.21 $\pm$ 5.84} \\ \hline \hline

SSDG-R \cite{jia2020single}       &  20.92          &  88.07          &  9.72                                                      &  22.57          &  85.61          &  15.95          &  21.75 $\pm$ 1.17          &  86.84 $\pm$ 1.74          &  12.84 $\pm$ 4.41           \\
\rowcolor[HTML]{EFEFEF}
 Ours: SSDG-R + DF-DM    &  \textbf{18.60} &  \textbf{88.80} &  \textbf{15.52}                                             &  \textbf{18.92} &  \textbf{88.59} &  \textbf{47.90} &  \textbf{18.76 $\pm$ 0.23} &  \textbf{88.70 $\pm$ 0.15} &  \textbf{31.71 $\pm$ 22.90}  \\ \hline \hline

 CDC \cite{yu2020searching}          &  28.94          &  78.96          &  13.93                                                      &  23.30          &  83.42          &  25.83           &  26.12 $\pm$ 3.99          &  81.19 $\pm$ 3.15          &  19.88 $\pm$ 8.41          \\
\rowcolor[HTML]{EFEFEF}
 Ous: CDC + DF-DM        &  \textbf{26.00} &  \textbf{81.73} &  \textbf{14.76}                                             &  \textbf{21.86} &  \textbf{85.77} &  \textbf{34.95}  &  \textbf{23.93 $\pm$ 2.93} &  \textbf{83.75 $\pm$ 2.86} &  \textbf{24.86 $\pm$ 14.28} \\ \hline
\end{tabular}%
\begin{tablenotes}
       \footnotesize
       \item $^*$ This paper adopts ResNet-18 as the backbone for CDC.
\end{tablenotes}
\end{threeparttable}
}
\end{table*}

In this case, MTCNN algorithm \cite{zhang2016joint} is adopted for face detection and alignment. All the detected faces are resized to (256,256). ResNet18 \cite{He_2016_CVPR} is set as the backbone for the feature extractor and discriminator. 


Besides the mentioned self-supervised methods, the state-of-the-art PA detectors, including DeepPixBiS \cite{george2019deep}, SSDG-R \cite{jia2020single}, CDC \cite{yu2020searching} are conducted in this paper. The effectiveness is validated by the improvement of such methods adopting the proposed method as initialization. And MS-LBP \cite{maatta2011face}, Binary CNN \cite{yang2014learn}, IDA \cite{wen2015face}, Color Texture \cite{boulkenafet2016face1}, LBP-TOP \cite{de2014face}, Auxiliary \cite{liu2018learning} and MADDG \cite{shao2019multi} are set as the baselines for reference.

This paper adopts the public platform pytorch for all experiments using  a workstation with CPUs of 2.8GHz, RAM of 512GB, and GPUs of NVIDIA Tesla V100.

\begin{table}[]
\centering
\caption{Performance Comparison between the Proposed Method and  Self-Supervised Methods in terms of EER(\%) $\downarrow$, AUC(\%) $\uparrow$ and TDR(\%)@FDR=1.0\% $\uparrow$ under the Cross-Material Setting on LivDet2017 and the Cross-Dataset Setting on [O, M] and [C, I] (Protocol-1). }
\label{tab:SSL}
\Huge
\resizebox{.48\textwidth}{!}{
\begin{threeparttable}
\begin{tabular}{c|ccc|ccc}
\hline
                         & \multicolumn{3}{c|}{Fingerprint}                & \multicolumn{3}{c}{Face}                       \\ \cline{2-7}
\multirow{-2}{*}{}       & EER(\%)           & AUC(\%)            & TDR(\%)            & EER(\%)            & AUC(\%)            & TDR(\%)          \\ \hline
Baseline                 & 13.26         & 93.46          & 29.28          & 42.30          & 59.51          & 4.14          \\ \hline
GAN based Discriminator \cite{isola2017image}  & 12.59         & 93.66          & 34.33          & 38.99          & 63.38          & 4.34          \\
AE based Encoder \cite{isola2017image}         & 11.20         & 94.83          & 27.78          & 35.65          & 65.79          & 2.70          \\ \hline
MoCo V2 \cite{chen2020improved}                 & 20.87         & 86.18          & 13.10          & 40.77          & 64.58          & 6.04          \\ \hline
\rowcolor[HTML]{EFEFEF}
Ours: DF-DM              & \textbf{8.87} & \textbf{96.55} & \textbf{56.42} & \textbf{31.28} & \textbf{73.68} & \textbf{10.06}         \\ \hline\hline
Pre-Trained from ImageNet & 4.10          & 99.06          & 73.92          & 26.90          & 79.10          & 11.37         \\ \hline
\rowcolor[HTML]{EFEFEF}
Ours:DF-DM (ImageNet)    & \textbf{2.59} & \textbf{99.55} & \textbf{90.96} & \textbf{18.78} & \textbf{89.62} & \textbf{30.39}         \\ \hline
\end{tabular}
\begin{tablenotes}
      \item  More details of the results in each case are given in the appendix.
\end{tablenotes}
\end{threeparttable}
}
\end{table}

\subsection{Effectiveness Analysis of the Proposed Method}
To quantify the contribution of De-Mixing and De-Folding, we test the performance of PAD with or without the corresponding pretext task. Table \ref{tab:ablation_finger} and Table \ref{tab:ablation_face} show the results on fingerprint and face cases, respectively.
And the corresponding ROC curves are shown in the appendix. In face PAD, we follow \cite{zhang2022effective} and use only protocol-1 for the ablation study. There are many more unseen PAs than the known ones in the training set. Using 3 datasets for training and 1 dataset for testing is not strict to the real application scenario. Hence, we use protocol-1 for our experiments. The numbers of data in O, C, I, and M are entirely different. To make the number of the training set and test set as close as possible, O and M are assigned to the same group. C and I are assigned to another group in protocol-1. Compared with protocol-2, protocol-1 uses fewer data for training, which makes the convergence of the network more difficult. It can show the generalization ability and effectiveness of our method.
The baseline is set as the model pre-trained from ImageNet for PAD. Compared with the baseline, both De-Folding and De-Mixing can provide more reasonable initialization. Specifically, an increase of 9.58\% in mean TDR@FDR=1.0\% is achieved by adopting De-Mixing as the pretext task in Table \ref{tab:ablation_finger}. When it comes to face, De-Folding improves the EER of baseline from 26.90\% to 22.86\%. This indicates that both components in the proposed method can promote PAD effectively. Among all the cases, the most significant improvement is obtained when all the designed components are adopted, i.e., DF-DM can reach 18.78\%, and 2.59\% mean EER in the face and fingerprint, respectively, which significantly outperforms those of baseline.

\begin{table*}[]
\centering
\caption{Performance of the Proposed Method with or without Each Component in terms of EER (\%) $\downarrow$, AUC (\%) $\uparrow$ and TDR(\%)@FDR=1.0\% $\uparrow$ under the Cross-Material Setting on LivDet2017.}
\label{tab:ablation_finger}
\Huge
\resizebox{1.\textwidth}{!}{
\begin{tabular}{ccc|ccccccccc|ccc}
\hline
                           &                                                                                 &                                                                                    & \multicolumn{3}{c|}{GreenBit}                                       & \multicolumn{3}{c|}{DigitalPersona}                                 & \multicolumn{3}{c|}{Oranthus}                   & \multicolumn{3}{c}{\textbf{Mean $\pm$ s.d.}}                                      \\ \cline{4-15}
\multirow{-2}{*}{Baseline} & \multirow{-2}{*}{\begin{tabular}[c]{@{}c@{}}In-Image\\ De-Folding\end{tabular}} & \multirow{-2}{*}{\begin{tabular}[c]{@{}c@{}}Out-of-Image\\ De-Mixing\end{tabular}} & EER(\%)            & AUC(\%)             & \multicolumn{1}{c|}{TDR(\%) } & EER(\%)            & AUC(\%)             & \multicolumn{1}{c|}{TDR(\%) } & EER(\%)            & AUC(\%)             & TDR(\%)   & EER(\%)                   & AUC(\%)                    & TDR(\%)          \\ \hline
$\surd$                    & $\times$                                                                        & $\times$                                                                           & 3.99          & 99.13          & 81.67                              & 5.08          & 98.80          & 68.64                              & 3.23          & 99.24          & 71.46          & 4.10 $\pm$ 0.93          & 99.06 $\pm$ 0.23          & 73.92 $\pm$ 6.86          \\ \hline
$\surd$                    & $\surd$                                                                         & $\times$                                                                           & 3.17          & 99.38          & 86.01                              & 3.90          & 99.03          & 72.78                              & 2.88          & 99.24          & 77.45          & 3.32 $\pm$ 0.53          & 99.22 $\pm$ 0.18          & 78.75 $\pm$ 6.71          \\ \hline
$\surd$                    & $\times$                                                                        & $\surd$                                                                            & 3.58          & 99.19          & 84.95                              & \textbf{3.44} & 99.14          & 74.56                              & 2.66          & 99.61          & 90.98          & 3.23 $\pm$ 0.50          & 99.31 $\pm$ 0.26          & 83.50 $\pm$ 8.31          \\ \hline
\rowcolor[HTML]{EFEFEF}
$\surd$                    & $\surd$                                                                         & $\surd$                                                                            & \textbf{2.88} & \textbf{99.65} & \textbf{94.14}                     & 3.49          & \textbf{99.26} & \textbf{81.36}                     & \textbf{1.40} & \textbf{99.75} & \textbf{97.37} & \textbf{2.59 $\pm$ 1.07} & \textbf{99.55 $\pm$ 0.26} & \textbf{90.96 $\pm$ 8.47} \\ \hline
\end{tabular}
}
\end{table*}

\begin{table*}[]
\centering
\caption{Performance of the Proposed Method with or without Each Component in terms of EER (\%) $\downarrow$, AUC (\%) $\uparrow$ and TDR(\%)@FDR=1.0\% $\uparrow$ under the Cross-Dataset Setting on OULU-NPU (O), CASIA-FASD (C), Idiap Replay-Attack (I) and MSU-MFSD (M)  (Protocol-1).}
\label{tab:ablation_face}
\setlength\tabcolsep{6pt}
\resizebox{1.\textwidth}{!}{
\begin{tabular}{ccc|cccccc|ccc}
\hline
                           &                                                                                 &                                                                                    & \multicolumn{3}{c|}{{[}O,M{]} to {[}C, I{]}}                         & \multicolumn{3}{c|}{{[}C,I{]} to {[}O, M{]}}     & \multicolumn{3}{c}{\textbf{Mean $\pm$ s.d.}}                                       \\ \cline{4-12}
\multirow{-2}{*}{Baseline} & \multirow{-2}{*}{\begin{tabular}[c]{@{}c@{}}In-Image\\ De-Folding\end{tabular}} & \multirow{-2}{*}{\begin{tabular}[c]{@{}c@{}}Out-of-Image\\ De-Mixing\end{tabular}} & EER(\%)            & AUC(\%)             & \multicolumn{1}{c|}{TDR(\%) } & EER(\%)             & AUC(\%)             & TDR(\%)   & EER(\%)                    & AUC(\%)                    & TDR(\%)          \\ \hline
$\surd$                    & $\times$                                                                        & $\times$                                                                           & 25.65          & 79.14          & 4.07                              & 28.14          & 79.05          & 18.66           & 26.90 $\pm$ 1.76          & 79.10 $\pm$ 0.06          & 11.37 $\pm$ 10.32         \\ \hline
$\surd$                    & $\surd$                                                                         & $\times$                                                                           & 20.33          & 84.51          & 9.79                              & 25.38          & 81.16          & 26.70           & 22.86 $\pm$ 3.57          & 82.84 $\pm$ 2.37          & 18.25 $\pm$ 11.96         \\ \hline
$\surd$                    & $\times$                                                                        & $\surd$                                                                            & 21.07          & 85.17          & 11.93                              & 26.33          & 80.25          & 27.49           & 23.70 $\pm$ 3.72          & 82.71 $\pm$ 3.48          & 19.71 $\pm$ 11.00         \\ \hline
\rowcolor[HTML]{EFEFEF}
$\surd$                    & $\surd$                                                                         & $\surd$                                                                            & \textbf{18.96} & \textbf{89.48} & \textbf{30.48}                     & \textbf{18.60} & \textbf{89.76} & \textbf{30.30} & \textbf{18.78 $\pm$ 0.25} & \textbf{89.62 $\pm$ 0.20} & \textbf{30.39 $\pm$ 0.13} \\ \hline
\end{tabular}%
}
\end{table*}

\section{Ablation Study}

\subsection{Comparison with Related Methods}
\subsubsection{Comparison with Self-Supervised Methods}
Due to the difference between the natural and face/fingerprint images, directly adopting existing self-supervised methods for PAD is not a proper choice.
Hence, we proposed a self-supervised learning-based method: DF-DM. The pipeline of self-supervised learning contains two training steps: self-supervised pretext task training and supervised downstream task training. In this paper, DF-DM is self-supervised pretext tasks, and the PAD classification is the downstream task. In the pretext tasks, we do not use any PAD labels, and the pseudo labels we use are automatically generated from the images themselves. We only use PAD labels in downstream tasks.
To validate the effectiveness of the proposed method, we compare DF-DM with the existing self-supervised methods. As the results listed in Table. \ref{tab:SSL}, the proposed method outperforms existing methods significantly. ``Ours: DF-DM" and ``Ours: DF-DM(ImageNet)" are training in self-supervised learning manner. ``Ours: DF-DM" denotes that in the pretext task training step (DF-DM), the network parameters are trained from scratch. But for ``Ours: DF-DM (ImageNet)", the network parameters are pre-train from ImageNet. ``Pre-Trained from ImageNet" is training in a supervised learning manner, so it does not contain the DF-DM training step. And in the supervised learning step (PAD classification), the network parameters are pre-train from ImageNet. In terms of face, when trained from scratch, our method can achieve an EER of 31.28\%, exceeding other self-supervised methods by around 4~10\% absolutely. Meanwhile, the proposed method can further improve the performance of the model pre-trained from ImageNet. Typically, in the case of fingerprint, DF-DM reaches 90.96\% TDR when FDR=1.0\%, which outperforms the initialization from ImageNet by a large margin, i.e., 90.96\% vs. 73.92\%. Note that the data scale of PAD is limited; hence, directly using MoCo cannot reach competitive results and may lead the model to learn useful features for identification, but it is useless for PAD.

\begin{table*}[!htb]
\centering
\caption{Performance Comparison between the Proposed Method and the State-Of-The-Art Methods on LivDet2017 under the Cross-Material and Cross-Sensor Settings in terms of Average Class Error (\%) $\downarrow$ and TDR(\%)@FDR=1.0\%$\uparrow$. }
\label{tab:finger_compare}
\setlength\tabcolsep{8pt}
\resizebox{1.\textwidth}{!}{
\begin{threeparttable}
\begin{tabular}{c|c|cccc|cc}
\hline
\multicolumn{2}{c|}{}                                                                                                                          & \multicolumn{2}{c}{Cross-Material Case}                                                              & \multicolumn{2}{c|}{Cross-Sensor Case}                                     & \multicolumn{2}{c}{\textbf{Mean}}                                     \\ \cline{3-8}
\multicolumn{2}{c|}{\multirow{-2}{*}{}}                                                                                                        & \multicolumn{1}{c}{ACE(\%)}                         & \multicolumn{1}{c}{TDR(\%)}                & \multicolumn{1}{c}{ACE(\%)}               & TDR(\%)                     & \multicolumn{1}{c}{ACE(\%)}     & TDR(\%)                \\ \hline
                                                                                                & LivDet 2017 Winner \cite{mura2018livdet}                          & 4.75 $\pm$ 1.40                                  & -                                                 & -                                      & -                                 & -                                 & -                                 \\
                                                                                                & F.S.B. \cite{8306930}                                      & 4.56 $\pm$ 1.12                                  & 73.32 $\pm$ 15.52                                 & 32.40$\pm$16.92                        & 21.26$\pm$28.06                   & 18.48                             & 50.69                             \\
\multirow{-3}{*}{Single Model}                                                                  & \cellcolor[HTML]{EFEFEF}Ours: DF-DM & \cellcolor[HTML]{EFEFEF}\textbf{2.48 $\pm$ 0.98} & \cellcolor[HTML]{EFEFEF} \textbf{90.96 $\pm$ 8.47} & \cellcolor[HTML]{EFEFEF}\textbf{19.82 $\pm$ 9.80}      & \cellcolor[HTML]{EFEFEF}\textbf{33.43 $\pm$ 24.12} & \cellcolor[HTML]{EFEFEF}\textbf{11.15} & \cellcolor[HTML]{EFEFEF}{\textbf{62.20}} \\\hline\hline
                                                                                                & F.S.B. + UMG Wrapper \cite{9079541}                        & \multicolumn{1}{c}{4.12 $\pm$ 1.34}             & \multicolumn{1}{c}{80.74 $\pm$ 10.02}            & \multicolumn{1}{c}{\textbf{20.37 $\pm$ 12.88}} & \textbf{43.23 $\pm$ 28.31}                 & 12.25                             & 61.99                            \\
\multirow{-2}{*}{\begin{tabular}[c]{@{}c@{}}Multiple Models\\ (Ensemble Learning)\end{tabular}} & RTK-PAD\cite{9457215}                                      & \multicolumn{1}{c}{\textbf{2.12 $\pm$ 0.72}}             & \multicolumn{1}{c}{\textbf{91.20 $\pm$ 7.59}}             & \multicolumn{1}{c}{21.87 $\pm$ 10.48} & 34.70 $\pm$ 25.30                 & \textbf{12.00}                             & \textbf{62.95}                             \\ \hline
\end{tabular}
\begin{tablenotes}
       \footnotesize
       \item The competing methods only report the result in ACE and TDR@FDR=1.0\%. We thus test the proposed method in ACE.
       \item More details of the results in each case are given in the appendix.
\end{tablenotes}
\end{threeparttable}
}
\end{table*}

\subsubsection{Comparison with Presentation Attack Detectors}
To further verify the effectiveness of the proposed method, we compare it with state-of-the-art methods. As the results listed in Table \ref{tab:finger_compare}, under the cross-material and cross-sensor settings, the proposed method can outperform other single model-based methods by a large margin. In the cross-sensor case, compared with FSB, a reduction of 12.58\% in average classification error (ACE) can be obtained by DF-DM. By comprehensively analyzing cross-material and -sensor protocols, DF-DM can promote PA detector to 11.15\% mean ACE, even exceeding the multiple model-based methods, which convincingly proves the advantage of DF-DM. Regarding the cross-material case, a 2.48\% of ACE can be derived by our proposed method, which outperforms FSB+UMG Wrapper (4.12\%) and is close to  RTK-PAD (2.28\%). More details of the cross-material and cross-sensor are shown in the appendix.

\begin{table*}[!htbp]
	\caption{Performance Comparison between the Proposed Method and the State-Of-The-Art Methods in the Terms of HTER(\%) $\downarrow$, AUC(\%) $\uparrow$ under the Cross-Dataset Setting of Face  (Protocol-2).  }
	\label{tab:face_protocol2}
\resizebox{.98\textwidth}{!}{
\begin{threeparttable}
\begin{tabular}{c|cc|cc|cc|cc|cc}
\hline
                         & \multicolumn{2}{c|}{{[}O, C, I{]} to M} & \multicolumn{2}{c|}{{[}O, M, I{]} to C} & \multicolumn{2}{c|}{{[}O, C, M{]} to I} & \multicolumn{2}{c|}{{[}I, C, M{]} to O} & \multicolumn{2}{c}{\textbf{Mean $\pm$ S.d.}}         \\ \cline{2-11}
\multirow{-2}{*}{Method} & HTER (\%)          & AUC (\%)           & HTER (\%)          & AUC (\%)           & HTER (\%)          & AUC (\%)           & HTER (\%)          & AUC (\%)           & HTER (\%)                 & AUC (\%)                  \\ \hline
MS-LBP \cite{maatta2011face}                   & 29.76              & 78.50              & 54.28              & 44.98              & 50.30              & 51.64              & 50.29              & 49.31              & 46.16 $\pm$ 9.61          & 56.11 $\pm$ 13.15         \\
Binary CNN \cite{yang2014learn}               & 29.25              & 82.87              & 34.88              & 71.94              & 34.47              & 65.88              & 29.61              & 77.54              & 32.05 $\pm$ 2.63          & 74.56 $\pm$ 6.33          \\
IDA \cite{wen2015face}                     & 66.67              & 27.86              & 55.17              & 39.05              & 28.35              & 78.25              & 54.20              & 44.59              & 51.10 $\pm$ 14.02         & 47.44 $\pm$ 18.78         \\
Color Texture \cite{boulkenafet2016face1}           & 28.09              & 78.47              & 30.58              & 76.89              & 40.40              & 62.78              & 63.59              & 32.71              & 40.67 $\pm$ 14.01         & 62.71 $\pm$ 18.37         \\
LBP-TOP \cite{de2014face}                  & 36.90              & 70.80              & 33.52              & 73.15              & 29.14              & 71.69              & 30.17              & 77.61              & 32.43 $\pm$ 3.05          & 73.31 $\pm$ 2.62          \\
Auxiliary \cite{liu2018learning}               & -                  & -                  & 28.40              & -                  & 27.60              & -                  & -                  & -                  & -                         & -                         \\
MADDG \cite{shao2019multi}                   & 17.69              & 88.06              & 24.50              & 84.51              & 22.19              & 84.99              & 27.89              & 80.02              & 23.07 $\pm$ 3.71          & 84.40 $\pm$ 2.87          \\
SSDG-M  \cite{jia2020single}                  & 16.67              & 90.47              & 23.11              & 85.45              & 18.21              & 94.61              & 25.17              & 81.83              & 20.79 $\pm$ 3.47          & 88.09 $\pm$ 4.86          \\
ANRL  \cite{liu2021adaptive}                  & 16.03              & 91.04              & 10.83              & 96.75              & 17.85              & 89.26              & 15.67              & 91.90              & 15.10 $\pm$ 2.60          & 92.24 $\pm$ 2.77          \\ 
EPCR  \cite{wang2021consistency}                  & 12.50             & 95.30              & 18.90              & 89.70              & 14.00              & 92.40              & 17.90              & 90.90              & 15.83 $\pm$ 2.65          & 90.08 $\pm$ 2.09          \\
PatchNet  \cite{wang2022patchnet}                  & 7.10              & 98.46              & 11.33              & 94.58              & 13.40              & 95.67              & 11.82              & 95.07              & 10.91 $\pm$ 2.33          & 95.95 $\pm$ 1.50          \\ \hline
Baseline                 & 13.10              & 92.76              & 16.44              & 91.25              & 24.58              & 79.50              & 22.31              & 85.65              & 19.11 $\pm$ 4.57          & 87.29 $\pm$ 5.22          \\
\rowcolor[HTML]{EFEFEF}
Ours: DF-DM              & \textbf{7.14}      & \textbf{97.09}     & \textbf{15.33}     & \textbf{91.41}     & \textbf{14.03}     & \textbf{94.30}     & \textbf{16.68}     & \textbf{91.85}     & \textbf{13.30 $\pm$ 3.68} & \textbf{93.66 $\pm$ 2.26} \\ \hline
SSDG-R \cite{jia2020single}                  & 7.38               & 97.17              & 10.44              & 95.94              & 11.71              & \textbf{96.59}     & 15.61              & 91.54              & 11.29 $\pm$ 2.95          & 95.31 $\pm$ 2.22          \\
\rowcolor[HTML]{EFEFEF}
Ours: SSDG-R + DF-DM           & \textbf{5.71}      & \textbf{98.84}     & \textbf{10.00}     & \textbf{96.29}     & \textbf{8.68}      & 96.25              & \textbf{13.55}     & \textbf{93.27}     & \textbf{9.49 $\pm$ 2.81}  & \textbf{96.16 $\pm$ 1.97} \\ \hline
\end{tabular}
\begin{tablenotes}
       \footnotesize
       \item The competing methods only report the result in Half Total Error Rate (HTER) and AUC, we thus test the proposed method in such evaluation metrics.
\end{tablenotes}
\end{threeparttable}
}
\end{table*}

When it comes to face, we first conduct our proposed method in a famous publicly-available benchmark to justify its effectiveness. As listed in Table. \ref{tab:face_protocol2}, DF-DM can reach to 13.30\% Half Total Error Rate (HTER) and 93.66\% AUC. Without any changes in learning objectives and network architectures, a baseline model, ResNet-18, can directly surpass most competing methods by adopting DF-DM as the initialization. 
When combined with SSDG-R, the proposed method can achieve the best PAD performance, which improves SSDG-R from 11.29\% HTER to 9.49\%.
EPCR \cite{wang2021consistency} and PatchNet \cite{wang2022patchnet} in Table. \ref{tab:face_protocol2} are self-supervised based method. Our proposed DF-DM can suppress EPCR \cite{wang2021consistency} around 2.53\% in HTER.
Meanwhile, we re-implement some famous PA detectors and investigate the improvement from DF-DM in different detectors. As listed in Table \ref{tab:face_compare}, DF-DM can facilitate detection performance by around 5$\sim$25\% in mean TDR@FDR=1.0\%. When DeepPixBiS is used as the detector, DF-DM can improve the AUC of PAD from 82.42\% to 89.48\%. The experimental results indicate that the proposed method is general and can be integrated with various PA detectors.

\subsection{Intra-Dataset Testing}
We also conduct experiments on Rose-Youtu \cite{li2018unsupervised,li2022one} and WMCA \cite{george2019biometric} for intra-dataset testing. WMCA dataset \cite{george2019biometric} contains a wide variety of 2D and 3D presentation attacks, with a total of 1679 video samples from 72 subjects. Rose-Youtu dataset \cite{li2018unsupervised,li2022one} covers a large variety of illumination conditions, camera models, and attack types, which consists of 4225 videos with 25 subjects in total. Attack Presentation Classification Error Rate (APCER), Bona Fide Presentation Classification Error Rate (BPCER), and Average Classification Error Rate (ACER) are used for intra-dataset testing.

\begin{table}[]
\centering
\caption{Performance Comparison between the proposed method and SSDG \cite{jia2020single} on the WMCA dataset in the terms of APCER(\%) $\downarrow$, BPCER(\%)$\downarrow$ and ACER(\%)$\downarrow$.}
\label{tab:wmca}
\setlength\tabcolsep{8pt}
\resizebox{.48\textwidth}{!}{%
\begin{tabular}{c|cccc}
\cline{1-4}
{ }               & \multicolumn{1}{c}{{ APCER(\%)}} & \multicolumn{1}{c}{{ BPCER(\%)}} & \multicolumn{1}{c}{{ ACER(\%)}} &  \\ \cline{1-4}
{ Baseline}       & { 9.57}                          & { 3.62}                          & { 6.60}                         &  \\ \cline{1-4}
{ DF-DM}          & { 5.22}                          & \textbf{ 0.45}                          & { 2.84}                         &  \\ \cline{1-4}
\rowcolor[HTML]{EFEFEF}
{ SSDG-R + DF-DM} & \textbf{ 0.87}                          & { 0.91}                          & \textbf{ 0.89}                         &  \\ \cline{1-4}
\end{tabular}%
}
\end{table}


\begin{table}[]
\centering
\caption{Performance Comparison between the proposed method and SSDG \cite{jia2020single} on the Rose-Youtu dataset. in the terms of APCER(\%) $\downarrow$, BPCER(\%)$\downarrow$ and ACER(\%)$\downarrow$.}
\label{tab:roseyoutu}
\setlength\tabcolsep{8pt}
\resizebox{.48\textwidth}{!}{%
\begin{tabular}{c|cccc}
\cline{1-4}
{ }          & { APCER(\%)} & { BPCER(\%)} & { ACER(\%)} &  \\ \cline{1-4}
{ Baseline}  & { 1.78}  & { 2.89}  & { 2.34} &  \\ \cline{1-4}
{ DF-DM}      & { 2.45}  & \textbf{ 1.17}  & { 1.81} &  \\ \cline{1-4}
\rowcolor[HTML]{EFEFEF}
{ SSDG-R + DF-DM} & \textbf{ 1.33}  & { 1.72}  & \textbf{ 1.53} &  \\ \cline{1-4}
\end{tabular}%
}
\end{table}

\begin{table*}[!b]
\centering
\caption{Performance comparison between different training strategies}
\label{tab:differenttrain}
\resizebox{.98\textwidth}{!}{%
\begin{tabular}{c|cc|cc|cc|cc|cc}
\hline
{ }                                                                               & \multicolumn{2}{c|}{{ [O, C, I] to M}}                   & \multicolumn{2}{c|}{{ [O, M, I] to C}}                    & \multicolumn{2}{c|}{{ [O, C, M] to I}}                    & \multicolumn{2}{c|}{{ [I, C, M] to O}}                    & \multicolumn{2}{c}{{ Mean ± S.d.}}                                      \\ \cline{2-11} 
\multirow{-2}{*}{{ \begin{tabular}[c]{@{}c@{}}Training \\ strategy\end{tabular}}} & { HTER(\%)}      & { AUC(\%)}        & { HTER(\%)}       & { AUC(\%)}        & { HTER(\%)}       & { AUC(\%)}        & { HTER(\%)}       & { AUC(\%)}        & { HTER(\%)}              & { AUC(\%)}               \\ \hline
{ Together training}                                                                  & { 8.57}          & { 96.20}          & { 20.67}          & { 88.37}          & { 21.00}          & { 81.90}          & { \textbf{16.35}} & { 91.35}          & { 16.65 ± 5.01}          & { 89.46 ± 5.18}          \\\rowcolor[HTML]{EFEFEF}
{ Step Training}                                                              & { \textbf{7.14}} & { \textbf{97.09}} & { \textbf{15.33}} & { \textbf{91.41}} & { \textbf{14.03}} & { \textbf{94.30}} & { 16.68}          & { \textbf{91.85}} & { \textbf{13.30 ± 3.68}} & { \textbf{93.66 ± 2.26}} \\ \hline
\end{tabular}%
}
\end{table*}

\subsubsection{Results on WMCA}
The WMCA dataset consists of 1941 short video recordings of both bonafide and presentation attacks from 72 identities. The data is recorded from several channels, including color, depth, infrared, and thermal. We randomly select one frame from each video for training and testing. Hence, the training and testing sets contain a total of 1941 images. The experiment results on Rose-Youtu are listed in Table \ref{tab:wmca}. DF-DM can achieve 2.84\% ACER, exceeding baseline performance by 3.76\%. The combination of SSDG-R with our method gets a better result, which improves the baseline from 6.60\% ACER to 0.89\% ACER.

\subsubsection{Results on Rose-Youtu}
The Rose-Youtu dataset consists of 4225 videos with 25 subjects in total. For each video, we use the same method as WMCA to extract images, and the dataset contains a total of 4255 images. The experiment results on Rose-Youtu are listed in Table \ref{tab:roseyoutu}. Our proposed DF-DM outperforms the baseline method by around 0.53\% in ACER. When combined with SSDG-R, the proposed method gets the best results.

\subsection{Training with different strategies}


The proposed method can be trained with different strategies, including step and together training. In the step training, it first uses the self-supervised learning task DF-DM to train the model without using PAD labels and then fine-tune the network parameters in a supervised training manner with PAD labels. In the together training, the two training steps: the DF-DM training without PAD labels and supervised training with PAD labels, are trained simultaneously. Experimental results are shown in Table. \ref{tab:differenttrain}, which indicates that step training outperforms together training by a wide margin.




\section{Visualization Results}
To further investigate the advance of the proposed method over other baselines, we visualize the discriminative features extracted by the models with the same architecture but different initialization. Besides, to further clarify the contribution of DF-DM for the generalization, we then visualize the discriminative features among different datasets. The visualization results are presented in the appendix.


\section{Conclusion}
In this paper, we proposed a self-supervised learning-based method to improve the generalization performance of PA detectors.
De-Folding and De-Mixing pretext tasks included in the method work together as a local-global strategy. That is, De-Folding requires the model to reconstruct the folded image to the row by extracting region-specific features, i.e., local information, while De-Mixing drives the model to derive instance-specific features, i.e., global information, by disentangling the mixed samples. The generalization ability is finally improved by the comprehensive local-global view of training samples. The effectiveness of the proposed method is verified in terms of face and fingerprint PAD, including 7 publicly-available datasets: LivDet2017, OULU-NPU, CASIA-FASD, Idiap Replay-Attack, MSU-MFSD, WMCA, and Rose-Youtu. In the future, we will further investigate the application of the proposed method in other tasks, such as fingerprint/face recognition and face detection/alignment.


{\small
\bibliographystyle{IEEEtran}
\bibliography{myreference}
}

\begin{IEEEbiography}[{\includegraphics[width=1in,height=1.25in,clip, keepaspectratio]{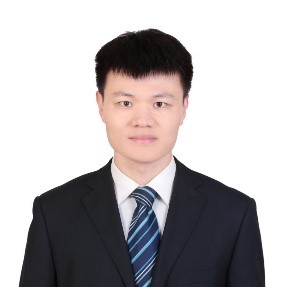}}]
{Zhe Kong} received the B.S. degree from the South China Agricultural University, in 2020. He is currently pursuing the M.S. degree with Shenzhen University. His research interests include image processing, face anti-spoofing and self-supervised learning.
\end{IEEEbiography}

\begin{IEEEbiography}[{\includegraphics[width=1in,height=1.25in,clip, keepaspectratio]{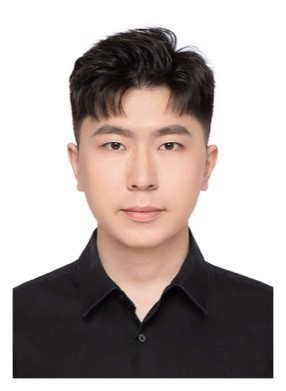}}]
{Wentian Zhang} received the B.S. degree from the Shaanxi University of Science and Technology, in 2020. He is currently pursuing the M.S. degree with Shenzhen University. His research interests include image processing, anomaly detection and graph embedding.
\end{IEEEbiography}

\begin{IEEEbiography}[{\includegraphics[width=1in,height=1.25in,clip, keepaspectratio]{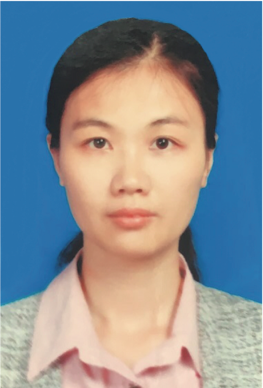}}]
{Feng Liu} is currently an Associate Professor at School of Computer Science \& Software Engineering, Shenzhen University. She obtained her B.Sc. and M.Sc. degrees both from Xidian University, Xi'an, Shaanxi, China. She received her Ph.D. degree in computer science from the Department of Computing at the Hong Kong Polytechnic University in 2014. Her research interests include pattern recognition and image processing, especially focus on their applications to fingerprints. Dr. Liu has published more than 40 papers in academic journals and conferences and participated in many research projects either as principal investigators or as primary researchers. She is a reviewer for many renowned field journals and conferences and a member of the IEEE.
\end{IEEEbiography}

\begin{IEEEbiography}[{\includegraphics[width=1in,height=1.25in,clip, keepaspectratio]{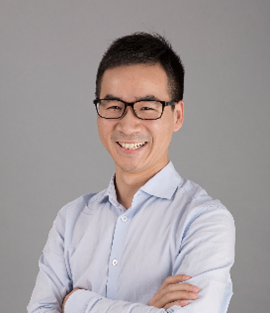}}]
{Wenhan Luo} is currently an Associate Professor with Sun Yat-sen University. Prior to that, he worked as a research scientist for Tencent and Amazon. He has published over 40 papers in top conferences and leading journals, including ICML, CVPR, ICCV, ECCV, ACL, AAAI, ICLR, TPAMI, IJCV, AI, TIP, etc. He also has been reviewer, senior PC member and Guest Editor for several prestigious journals and conferences. His research interests include several topics in computer vision and machine learning, such as image/video synthesis, image/video quality restoration, reinforcement learning. He received the Ph.D. degree from Imperial College London, UK, 2016, M.E. degree from Institute of Automation, Chinese Academy of Sciences, China, 2012 and B.E. degree from Huazhong University of Science and Technology, China, 2009.
\end{IEEEbiography}

\begin{IEEEbiography}[{\includegraphics[width=1in,height=1.25in,clip, keepaspectratio]{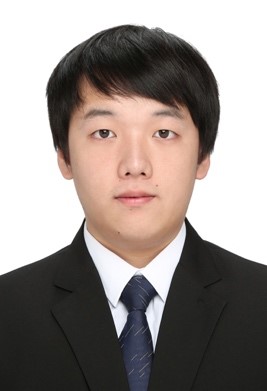}}]
{Haozhe Liu} is a PhD student at AI Initiative, King Abdullah University of Science and Technology (KAUST). He received the M.S. degree in Computer Science from Shenzhen University in 2022.  He had co-authored several papers on top-tier journals and conferences, including CVPR, ICCV, ECCV, MICCAI, IEEE trans on Image Processing, etc.. He serves as a reviewer of the top-tier conferences, e.g. CVPR’2022, ICML’2022, ECCV’2022 and MICCAI’2022. His research interests include regularization, self-supervised learning, adversarial learning and reinforcement learning.
\end{IEEEbiography}

\begin{IEEEbiography}[{\includegraphics[width=1in,height=1.25in,clip, keepaspectratio]{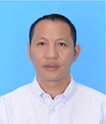}}]
{Linlin Shen} received the B.Sc. and M.Eng. degrees from Shanghai Jiaotong University, Shanghai, China, and the Ph.D. degree from the University of Nottingham, Nottingham, U.K. He was a Research Fellow with the University of Nottingham, working on MRI brain image processing. He is currently a Pengcheng Scholar Distinguished Professor with the School of Computer Science and Software Engineering, Shenzhen University, Shenzhen, China. He is also the Honorary Professor with the School of Computer Science, University of Nottingham, U.K., and a Distinguished Visiting Scholar with the University of Macao, China. He serves as the Director of the Computer Vision Institute, AI Research Center for Medical Image Analysis and Diagnosis and China-U.K., joint Research Lab for Visual Information Processing. He is listed as the Most Cited Chinese Researchers by Elsevier. His research interests include deep learning, facial recognition, analysis/synthesis and medical image processing. He received the Most Cited Paper Award from the Journal of Image and Vision Computing. His cell classiﬁcation algorithms were the winners of the International Contest on Pattern Recognition Techniques for Indirect Immunoﬂuorescence Images held by ICIP 2013 and ICPR 2016.
\end{IEEEbiography}

\begin{IEEEbiography}[{\includegraphics[width=1in,height=1.25in,clip, keepaspectratio]{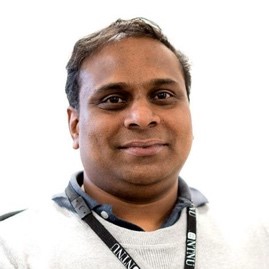}}]
{Raghavendra Ramachandra} obtained a Ph.D. in computer science and technology from the University of Mysore, Mysore India and Institute Telecom, and Telecom Sudparis, Evry, France (carried out as collaborative work) in 2010. He is currently a full professor at the Institute of Information Security and Communication Technology (IIK), Norwegian University of Science and Technology (NTNU), Gjøvik, Norway. He was a researcher with the Istituto Italiano di Tecnologia, Genoa, Italy, where he worked with video surveillance and social signal processing. His main research interests include deep learning, machine learning, data fusion schemes, and image/video processing, with applications to biometrics, multi-modal biometric fusion, human behaviour analysis, and crowd behaviour analysis. He has authored several papers and is a reviewer for several international conferences and journals. He also holds several patents in biometric presentation attack detection and morphing attack detection. He has also been involved in various conference organising and program committees and has served as an associate editor for various journals. He has participated (as a PI, co-PI or contributor) in several EU projects, IARPA USA and other national projects. He is serving as an editor of the ISO/IEC 24722 standards on multi-modal biometrics and an active contributor to the ISO/IEC SC 37 standards on biometrics. He has received several best paper awards, and he is also a senior member of IEEE.
\end{IEEEbiography}

\end{document}


\title{Taming Self-Supervised Learning for Presentation Attack Detection: De-Folding and De-Mixing}
\captionsetup[table]{labelfont={bf},name={TABLE A.},labelsep=period}
\captionsetup[figure]{labelfont={bf},name={Figure A.},labelsep=period}
\maketitle
\thispagestyle{empty}
\section{Details of the Datasets}
To clarify our experimental settings, we give the details of the datasets in this section. Table A.\ref{tab:dataset_f} summarizes the information of LivDet2017, which is used to evaluate the performance of fingerprint PAD. LivDet 2017 consists of over 17500 fingerprint images, captured from GreenBit, Orcanthus and Digital Persona. Specifically, GreenBit is employed for Italian border controls and issurance of Italian electronic documents. Orcanthus is widely used in the personal computer (PC). And Digital Persona is adopted in the mobile devices, such as Nexus 7 tablet. Hence, the adopted readers are reliable to test the practical performance for fingerprint PAD. More details are available at Table A.\ref{tab:dataset_f}.
\begin{table*}[!htbp]
\setlength\tabcolsep{28pt}
\caption{ Details of LivDet2017 Datasets.}
\label{tab:dataset_f}
\resizebox{.98\textwidth}{!}{%
\begin{threeparttable}
\begin{tabular}{c|ccc}
\hline
\multirow{2}{*}{Fingerprint Reader}       & \multicolumn{1}{c|}{GreenBit}     & \multicolumn{1}{c|}{Orcanthus}     & Digital Persona \\ 
                   & \multicolumn{1}{c|}{DactyScan84c} & \multicolumn{1}{c|}{Certis2 Imag}  & U.are.U 5160    \\ \hline
Type                     & \multicolumn{1}{c|}{Optical}      & \multicolumn{1}{c|}{Thermal swipe} & Optical         \\ \hline
Application                     & \multicolumn{1}{c|}{Italian Border Control}      & \multicolumn{1}{c|}{Personal Computer} & Nexus 7 Tablet         \\ \hline
Image Size(w$\times$h)          & \multicolumn{1}{c|}{500 $\times$ 500}    & \multicolumn{1}{c|}{-$^*$}       & 252 $\times$ 324       \\ \hline
Resolution(dpi)          & \multicolumn{1}{c|}{500}          & \multicolumn{1}{c|}{500}           & 500             \\ \hline
Live Image(Trian/Test)   & \multicolumn{1}{c|}{1000/1700}    & \multicolumn{1}{c|}{1000/1700}     & 999/1700        \\ \hline
Spoof Images(Train/Test) & \multicolumn{1}{c|}{1200/2040}    & \multicolumn{1}{c|}{1200/2018}     & 1199/2028       \\ \hline
Spoof Materials for Training   & \multicolumn{3}{c}{Wood Glue, Exoflex,   Body Double}                                    \\ \hline
Spoof Materials for Testing  & \multicolumn{3}{c}{Gelatine, Liquid Ecoflex, Latex}                                    \\ \hline
\end{tabular}%
\begin{tablenotes}
       \footnotesize
       \item $^*$ The images derived from Orcanthus reader are with the variable size.
\end{tablenotes}
\end{threeparttable}
}
\end{table*}

Meanwhile, we conclude the information for the datasets with respect to face PAD. As listed in Table A. \ref{tab:dataset_face}, four datasets, including Oulu-NPU, CASIA-FASD, Idiap Replay-Attack and MSU-MFSD, are used to evaluate the performance of PAD in this paper. In particular, printed photos, display photos and replayed videos are adopted as the attacks for facial recognition system. Various acquisition devices, such as laptop and smartphone, are considered in different datasets. Hence the robustness and generalization of the PAD methods can be tested through cross-dataset protocol. In such protocol, [O,M] and [C,I] are set as two different groups. The model is trained on a given group and tested on the other one. For each video, only one randomly-selected frame is used to train or test the detectors. 
\begin{table*}[!htbp]
\setlength\tabcolsep{16pt}
\caption{Summary of Oulu-NPU, CASIA-FASD, Idiap Replay-Attack and MSU-MFSD.}
\label{tab:dataset_face}
\large
\resizebox{.98\textwidth}{!}{%
\begin{tabular}{c|c|c|c|c|c|c|c}
\hline
                    & \multirow{2}{*}{Attack type}                                                           & \multicolumn{2}{c|}{Subject} & \multicolumn{2}{c|}{videos} & \multirow{2}{*}{Acquisition devices}                            & \multirow{2}{*}{Display devices}                                     \\ \cline{3-6}
                    &                                                                                        & train         & test         & real        & attack        &                                                                 &                                                                      \\ \hline
Oulu-NPU (O)           & \begin{tabular}[c]{@{}c@{}}Printed photo\\ Display photo\\ Replayed video\end{tabular} & 35            & 20           & 1980        & 3960          & 6 smartphone                                                    & \begin{tabular}[c]{@{}c@{}}Dell 1950FP\\ Macbook Retina\end{tabular} \\ \hline
CASIA-FASD (C)         & \begin{tabular}[c]{@{}c@{}}Printed photo\\ Cut photo\\ Replayed video\end{tabular}     & 20            & 30           & 150         & 450           & 3 webcams                                                       & iPad                                                                 \\ \hline
Idiap Replay-Attack (I)& \begin{tabular}[c]{@{}c@{}}Printed photo\\ Replay video\end{tabular}                   & 30            & 20           & 390         & 640           & 1 laptop                                                        & \begin{tabular}[c]{@{}c@{}}iPhone 3GS\\ iPad\end{tabular}            \\ \hline
MSU-MFSD (M)            & \begin{tabular}[c]{@{}c@{}}Printed photo\\ Cut photo\\ Replayed video\end{tabular}     & 18            & 17           & 110         & 330           & \begin{tabular}[c]{@{}c@{}}1 laptop\\ 1 smartphone\end{tabular} & \begin{tabular}[c]{@{}c@{}}iPad Air\\ iphone 5s\end{tabular}         \\ \hline
\end{tabular}%
}
\end{table*}

\section{Ablation Study}
\begin{figure*}
  \centering
  \includegraphics[width=.98\textwidth]{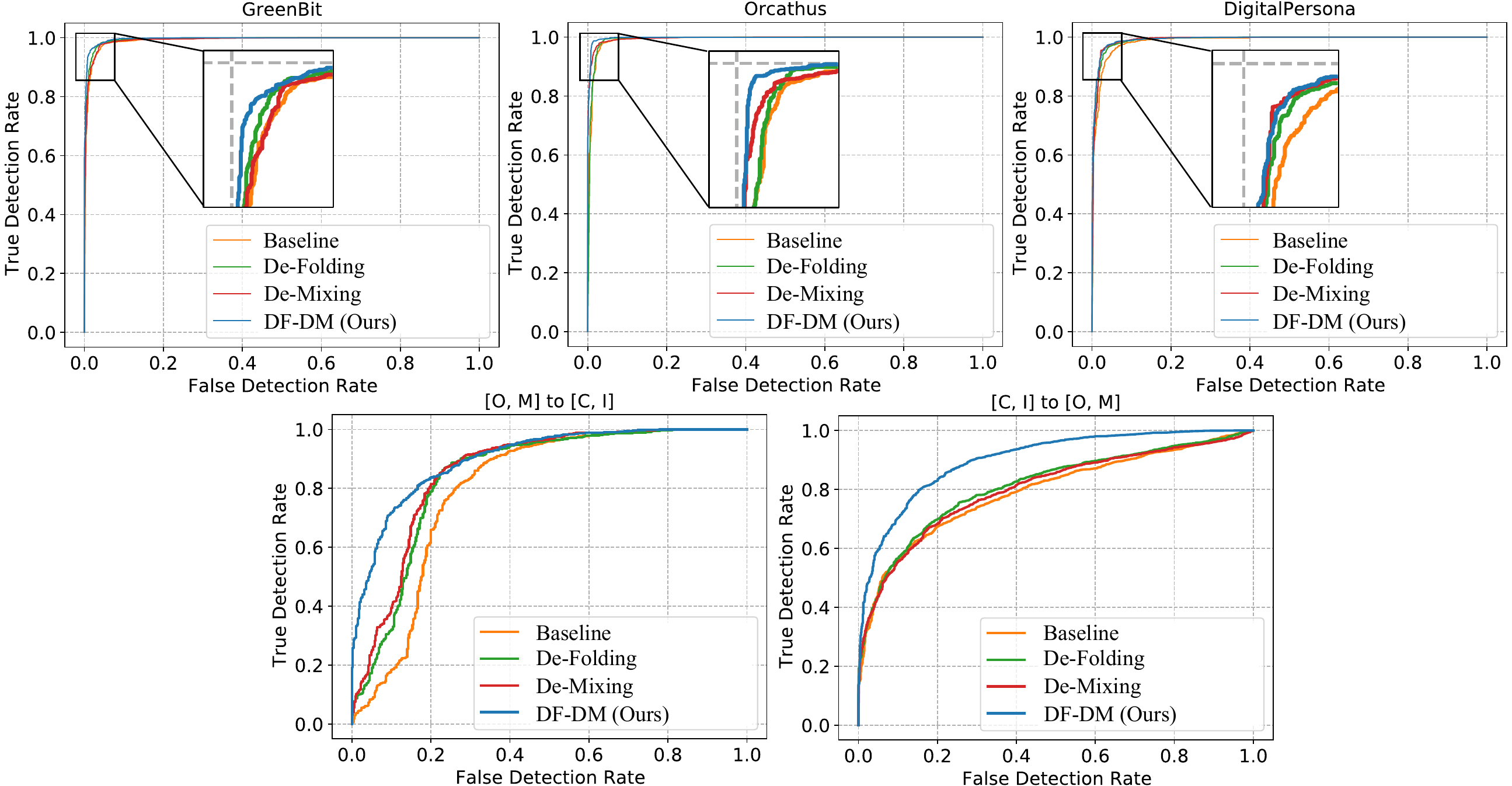}
  \caption{ROC curves for the ablation study. The first row presents the cases on LivDet2017 under the cross-material setting. The second row refers to the face anti-spoofing under the cross-dataset setting on [O,M] and [C,I].   
  }
  \label{fig:roc_ablation}
\end{figure*}

To quantify the contribution of De-Mixing and De-Folding, we test the discriminative performance of the variants with or without each pretext tasks. Fig. A.\ref{fig:roc_ablation} shows the ROC curves carried on fingerprint and face cases respectively, which is corresponding to Table 1 and Table 2 in the main body. As shown in Fig. A.\ref{fig:roc_ablation}, the first row refers to the fingerprint case on LivDet2017 and the second row is the face case on Oulu-NPU, CASIA-FASD, Idiap and MSU-MFSD. The proposed method with blue line shows a distinct advance over the other baselines among all the cases.
\begin{table*}[!hbp]
\centering
\caption{ Performance Comparison between the Proposed Method and the Self-Supervised Methods on LivDet2017 under the Cross-Material Setting in the Terms of EER(\%)$\downarrow$, AUC(\%)$\uparrow$ and TDR(\%)@FDR=1.0\%$\uparrow$.}
\Huge
\label{tab:fingerS}
\resizebox{\textwidth}{!}{%
\begin{tabular}{c|ccc|ccc|ccc|ccc}
\hline
                                  & \multicolumn{3}{c|}{GreenBit}                   & \multicolumn{3}{c|}{DigitalPersona}             & \multicolumn{3}{c|}{Oranthus}                   & \multicolumn{3}{c}{\textbf{Mean $\pm$ s.d.}}                                       \\ \cline{2-13} 
                                  & EER(\%)           & AUC(\%)            & TDR(\%)  & EER(\%)           & AUC(\%)            & TDR(\%)  & EER(\%)           & AUC(\%)            & TDR(\%)  & EER(\%)                  & AUC(\%)                   & TDR(\%)          \\ \hline
Baseline                 & 10.66         & 96.20          & 41.97          & 15.24         & 92.12          & 40.43          & 13.88         & 92.06          & 5.45           & 13.26 $\pm$ 2.35         & 93.46 $\pm$ 2.37          & 29.28 $\pm$ 20.65          \\
GAN based Discriminator  & 9.33          & 96.03          & 40.86          & 12.34         & 94.07          & 51.33          & 16.11         & 90.88          & 10.80          & 12.59 $\pm$ 3.40         & 93.66 $\pm$ 2.60          & 34.33 $\pm$ 21.04          \\
AE based Encoder         & 7.98          & 97.31          & 47.27          & 12.29         & 94.64 & 31.51          & 13.34         & 92.55          & 4.56           & 11.20 $\pm$ 2.84         & 94.83 $\pm$ 3.39          & 27.78 $\pm$ 21.60          \\
MoCo V2                  & 18.39         & 88.67          & 14.15          & 24.41         & 83.67          & 17.03          & 19.82         & 86.20          & 8.13           & 20.87 $\pm$ 3.15         & 86.18 $\pm$ 2.50          & 13.10 $\pm$ 4.54           \\
\rowcolor[HTML]{EFEFEF} 
Ours: DF-DM              & \textbf{7.68} & \textbf{97.61} & \textbf{59.14} & \textbf{9.27} & \textbf{96.54} & \textbf{68.20} & \textbf{9.66} & \textbf{95.51} & \textbf{41.92} & \textbf{8.87 $\pm$ 1.05} & \textbf{96.55 $\pm$ 1.05} & \textbf{56.42 $\pm$ 13.35} \\ \hline \hline
Pre-Trained from ImageNet & 3.99          & 99.13          & 81.67          & 5.08          & 98.80          & 68.64          & 3.23          & 99.24          & 71.46          & 4.10 $\pm$ 0.93          & 99.06 $\pm$ 0.23          & 73.92 $\pm$ 6.86           \\
\rowcolor[HTML]{EFEFEF} 
Ours: DF-DM (ImageNet)            & \textbf{2.88} & \textbf{99.65} & \textbf{94.14} & \textbf{3.49} & \textbf{99.26} & \textbf{81.36} & \textbf{1.40} & \textbf{99.75} & \textbf{97.37} & \textbf{2.59 $\pm$ 1.07} & \textbf{99.55 $\pm$ 0.26} & \textbf{90.96 $\pm$ 8.47}  \\ \hline
\end{tabular}%
}
\end{table*}
\begin{table*}[!htbp]
\centering
\large
\caption{Performance Comparison between the Proposed Method and the Self-Supervised Methods under the Cross-Dataset Setting on Oulu-NPU (O), CASIA-FASD (C), Idiap Replay-Attack(I) and MSU-MFSD (M) in the Terms of EER(\%)$\downarrow$, AUC(\%)$\uparrow$ and TDR(\%)@FDR=1.0\%$\uparrow$.}
\label{tab:faces}
\setlength\tabcolsep{10pt}
\resizebox{\textwidth}{!}{%
\begin{tabular}{c|cccccc|ccc}
\hline
                                 & \multicolumn{3}{c|}{{[}O,M{]} to {[}C, I{]}}                         & \multicolumn{3}{c|}{{[}C,I{]} to {[}O, M{]}}     & \multicolumn{3}{c}{\textbf{Mean $\pm$ s.d.}}                                       \\ \cline{2-10} 
                                 & EER(\%)            & AUC(\%)            & \multicolumn{1}{c|}{TDR(\%)} & EER(\%)            & AUC(\%)            &  TDR(\%) & EER(\%)                   & AUC(\%)                   &   TDR(\%)       \\ \hline
Baseline                & 37.50          & 64.65          & 2.21                               & 47.10          & 54.36          & 6.07           & 42.30 $\pm$ 6.79          & 59.51 $\pm$ 7.28          & 4.14 $\pm$ 2.73          \\ \hline
Gan based Discriminator & 36.39          & 65.02          & 0.14                               & 41.59          & 61.73          & 8.54           & 38.99 $\pm$ 3.68          & 63.38 $\pm$ 2.33          & 4.34 $\pm$ 5.94          \\
AE based Encoder        & 32.25          & 67.72          & 0.34                               & 39.05          & 63.86          & 5.06           & 35.65 $\pm$ 4.81          & 65.79 $\pm$ 2.73          & 2.70 $\pm$ 3.34          \\
MoCo V2                 & 33.75          & 68.00          & 0.00                               & 47.79          & 61.16          & 12.07           & 40.77 $\pm$ 9.93          & 64.58 $\pm$ 4.84          & 6.04 $\pm$ 8.53          \\ \hline
\rowcolor[HTML]{EFEFEF} 
Ours: DF-DM             & \textbf{29.61} & \textbf{73.78} & \textbf{6.76}                     & \textbf{32.94} & \textbf{73.58} & \textbf{13.36}  & \textbf{31.28 $\pm$ 2.35} & \textbf{73.68 $\pm$ 0.14} & \textbf{10.06 $\pm$ 4.67} \\ \hline \hline
Pre-Trained from ImageNet               & 25.65          & 79.14          & 4.07                              & 28.14          & 79.05          & 18.66           & 26.90 $\pm$ 1.76          & 79.10 $\pm$ 0.06          & 11.37 $\pm$ 10.32         \\ \hline
\rowcolor[HTML]{EFEFEF} 
Ours: DF-DM(ImageNet)            & \textbf{18.96} & \textbf{89.48} & \textbf{30.48}                     & \textbf{18.60} & \textbf{89.76} & \textbf{30.30} & \textbf{18.78 $\pm$ 0.25} & \textbf{89.62 $\pm$ 0.20} & \textbf{30.39 $\pm$ 0.13} \\ \hline
\end{tabular}
}
\end{table*}
\section{Comparison with Self-Supervised Methods }
\begin{figure*}[!hbtp]
  \centering
  \includegraphics[width=.98\textwidth]{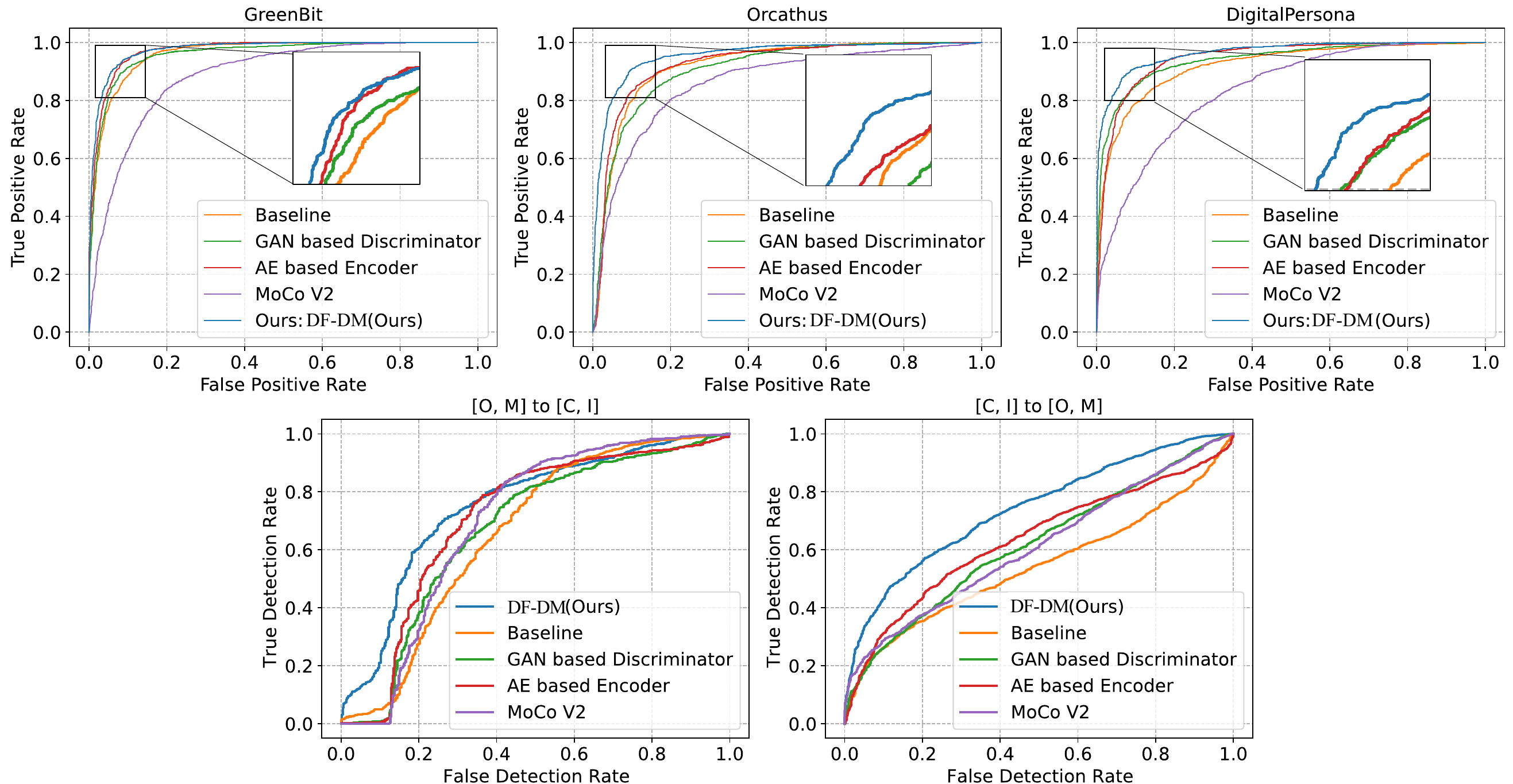}
  \caption{ROC curves for the proposed method and other self-supervised methods. The first row presents the cases on LivDet2017 under the cross-material setting. The second row refers to the face anti-spoofing under the cross-dataset setting on [O,M] and [C,I]. 
  }
  \label{fig:roc_self_supervised}
\end{figure*}

To prove the superiority of the proposed method, we compare DF-DM with other self-supervised learning based methods, including GAN based Discriminator, AE based Encoder and MoCo V2. The averaging performance is presented in the Table. 3 (see the main body). In this section, we will show the performance of each case under the cross-material and cross-dataset settings. 
Table A.\ref{tab:fingerS} presents the performance carried on LivDet2017 in the terms of Equal Error Rate (EER), Area Under Curve (AUC) and True Detection Rate (TDR) when False Detection Rate (FDR)=1.0\%. Compared with the competing self-supervised methods, the proposed method achieves the best results among all the readers and metrics, which convincingly proves the effectiveness of DF-DM. The most significant improvement is carried on the case of Orcanthus, where DF-DM facilitates the baseline(ImageNet) from 71.46 \% to  97.37 \% in  TDR@FDR=1.0\%.
When comes to the cross-dataset setting on face PAD, two cases, including [O,M] $\rightarrow$ [C,I] and [C,I] $\rightarrow$ [O,M] are considered. As listed in Table A.\ref{tab:faces}, the similar improvement can be achieved by the proposed method.
Typically, DF-DM can reach 73.58\% AUC in the case of [C,I] $\rightarrow$ [O,M], which significantly outperforms other solutions by a wide margin. Such results indicates the superiority of the proposed method over other self-supervised methods.  The ROC curves with more detailed information is presented in Fig. A.\ref{fig:roc_self_supervised}.

\begin{table*}[b]
\centering
\caption{Performance Comparison between the Proposed Method and the State-Of-The-Art Methods on LivDet2017 under the Cross-Material setting in the Terms of Average Class Error (\%)$\downarrow$ and TDR@FDR=1.0\%$\uparrow$.}
\label{tab:cross_material}
\Huge
\resizebox{\textwidth}{!}{%
\begin{tabular}{c|clcccc|cccc}
\hline
                              & \multicolumn{6}{c|}{Single Model}                                                                                                                                                & \multicolumn{4}{c}{Multiple Models (Ensemble Learning)}                                             \\ \cline{2-11} 
                              & \multicolumn{2}{c|}{LivDet 2017 Winner} & \multicolumn{2}{c|}{F.S.B}                           & \multicolumn{2}{c|}{\cellcolor[HTML]{EFEFEF}DF-DM(Ours)}                        & \multicolumn{2}{c|}{FSB. + UMG Wrapper} & \multicolumn{2}{c}{RTK-PAD}     \\ \cline{2-11} 
\multirow{-3}{*}{LivDet 2017} & \multicolumn{2}{c|}{ACE(\%)}            & ACE(\%)     & \multicolumn{1}{c|}{TDR(\%)} & \cellcolor[HTML]{EFEFEF}ACE(\%)     & \cellcolor[HTML]{EFEFEF}TDR(\%) & ACE(\%)            & \multicolumn{1}{c|}{TDR(\%)}       & ACE(\%)     &TDR(\%) \\ \hline
GreenBit                      & \multicolumn{2}{c}{3.56}                & 3.32        & 91.07                                  & \cellcolor[HTML]{EFEFEF}\textbf{2.78}        & \cellcolor[HTML]{EFEFEF}\textbf{94.14}             & 2.58               & 92.29                                        & \textbf{1.92}        & \textbf{96.82}             \\
Digital Persona               & \multicolumn{2}{c}{6.29}                & 4.88        & 62.29                                  & \cellcolor[HTML]{EFEFEF}\textbf{3.28}        & \cellcolor[HTML]{EFEFEF}\textbf{81.36}             & 4.80                & 75.47                                        & \textbf{3.25}        & \textbf{80.57}             \\
Orcanthus                     & \multicolumn{2}{c}{4.41}                & 5.49        & 66.59                                  & \cellcolor[HTML]{EFEFEF}\textbf{1.38}        & \cellcolor[HTML]{EFEFEF}\textbf{97.37}             & 4.99               & 74.45                                        & \textbf{1.67}        & \textbf{96.18}             \\ \hline
\textbf{Mean $\pm$ s.d.}                   & \multicolumn{2}{c}{4.75 $\pm$ 1.40}         & 4.56 $\pm$ 1.12 & 73.32 $\pm$ 15.52                          & \cellcolor[HTML]{EFEFEF}\textbf{2.48 $\pm$ 0.98} & \cellcolor[HTML]{EFEFEF}\textbf{90.96 $\pm$ 8.47}      & 4.12 $\pm$ 1.34        & 80.74 $\pm$ 10.02                                & \textbf{2.28 $\pm$ 0.69} & \textbf{91.19 $\pm$ 7.51}      \\ \hline
\end{tabular}%
}
\end{table*}

\section{Comparison with the State-Of-The-Art Detector in Fingerprint}
To further verify the effectiveness of the proposed method, we compare DF-DM with the state-of-the-art detectors on LivDet2017 under the cross-material and cross-sensor settings. As listed in Table A.\ref{tab:cross_material} and Table A.\ref{tab:cross_sensor}, the proposed method has the great improvement compared with other single model based methods. A typical case is that, in the cross-sensor protocol, DF-DM can achieve 33.43\% TDR@FDR=1.0\%, while FSB only reaches 21.26\%. Note that even compared with multiple model based methods, DF-DM also achieves very competitive results. In the terms of cross-material case, a 2.48\% of ACE can be derived by our proposed method, which outperforms FSB+UMG Wrapper (4.12\%) and  is close to  RTK-PAD (2.28\%). 

\begin{table}[]
\centering
\caption{Performance Comparison between the Proposed Method and the State-Of-The-Art Methods on LivDet2017 under the Cross-Sensor setting in the Terms of Average Class Error (\%)$\downarrow$ and TDR@FDR=1.0\%$\uparrow$.}
\label{tab:cross_sensor}
\Huge
\resizebox{1.\textwidth}{!}{
\begin{tabular}{c|cc|cc|cccc}
\hline
                                             & \multicolumn{4}{c|}{Single Model}                                                                                                              & \multicolumn{4}{c}{Multiple Models (Ensemble Learning)}                                               \\ \cline{2-9} 
                                             & \multicolumn{2}{c|}{FSB}              & \multicolumn{2}{c|}{\cellcolor[HTML]{EFEFEF}Ours:DF-DM}                                                & \multicolumn{2}{c|}{RTK-PAD}                 & \multicolumn{2}{c}{FSB+UMG}                            \\ \cline{2-9} 
\multirow{-3}{*}{Training (Testing) Readers} & ACE(\%)               & TDR(\%)               & \cellcolor[HTML]{EFEFEF}ACE(\%)                       & \cellcolor[HTML]{EFEFEF}TDR(\%)                        & ACE(\%)               & \multicolumn{1}{c|}{TDR(\%)} & ACE(\%)                        & TDR(\%)                        \\ \hline
GreenBit (Orcanthus)                         & 50.57             & 0.00              & \cellcolor[HTML]{EFEFEF}\textbf{19.42}            & \cellcolor[HTML]{EFEFEF}\textbf{25.87}             & \textbf{30.49}    & 20.61                    & 33.95                      & \textbf{21.52}             \\ \cline{1-1}
GreenBit (Digital Persona)                   & 10.63             & 57.48             & \cellcolor[HTML]{EFEFEF}\textbf{8.46}             & \cellcolor[HTML]{EFEFEF}\textbf{72.14}             & 7.41              & 70.41                    & \textbf{5.19}              & \textbf{72.91}             \\ \cline{1-1}
Orcanthus (GreenBit)                         & 30.07             & 8.02              & \cellcolor[HTML]{EFEFEF}\textbf{29.87}            & \cellcolor[HTML]{EFEFEF}\textbf{20.25}             & 28.81             & 15.00                    & \textbf{18.25}             & \textbf{30.91}             \\ \cline{1-1}
Orcanthus (Digital Persona)                  & 42.01             & 4.97              & \cellcolor[HTML]{EFEFEF}\textbf{27.21}            & \cellcolor[HTML]{EFEFEF}\textbf{7.94}              & 29.19             & 13.26                    & \textbf{23.64}             & \textbf{28.46}             \\ \cline{1-1}
Digital Persona (GreenBit)                   & 10.46             & \textbf{57.06}    & \cellcolor[HTML]{EFEFEF}\textbf{7.49}             & \cellcolor[HTML]{EFEFEF}53.03                      & 6.74              & 70.25                    & \textbf{3.65}              & \textbf{85.21}             \\ \cline{1-1}
Digital Persona (Orcanthus)                  & 50.68             & 0.00              & \cellcolor[HTML]{EFEFEF}\textbf{26.44}            & \cellcolor[HTML]{EFEFEF}\textbf{21.36}             & \textbf{28.55}    & 18.68                    & 31.56                      & \textbf{20.38}             \\ \hline
Mean $\pm$ s.d.                              & 32.40 $\pm$ 16.92 & 21.26 $\pm$ 28.06 & \cellcolor[HTML]{EFEFEF}\textbf{19.82 $\pm$ 9.80} & \cellcolor[HTML]{EFEFEF}\textbf{33.43 $\pm$ 24.12} & 21.87 $\pm$ 10.48 & 34.70 $\pm$ 25.30        & \textbf{19.37 $\pm$ 12.88} & \textbf{43.23 $\pm$ 28.31} \\ \hline
\end{tabular}
}
\end{table}

\begin{figure*}[!htbp]
  \centering
  \includegraphics[width=.98\textwidth]{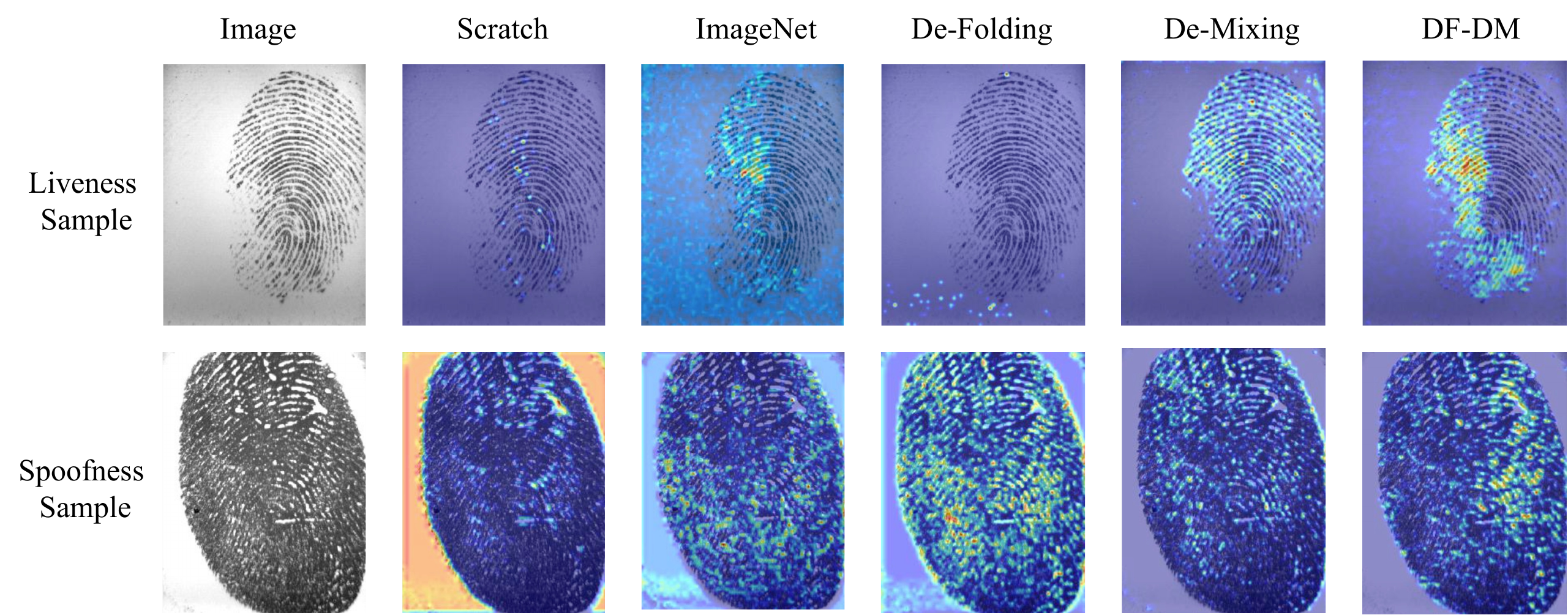}
  \caption{The Grad-CAM based visualization on LivDet2017 using Digital Persona. The first row shows the liveness sample and the second row presents the spoofness sample. Note that the visualized model is not trained by PAD task.  
  }
  \label{fig:vis_finger}
\end{figure*}
\begin{figure*}[!htbp]
  \centering
  \includegraphics[width=.98\textwidth]{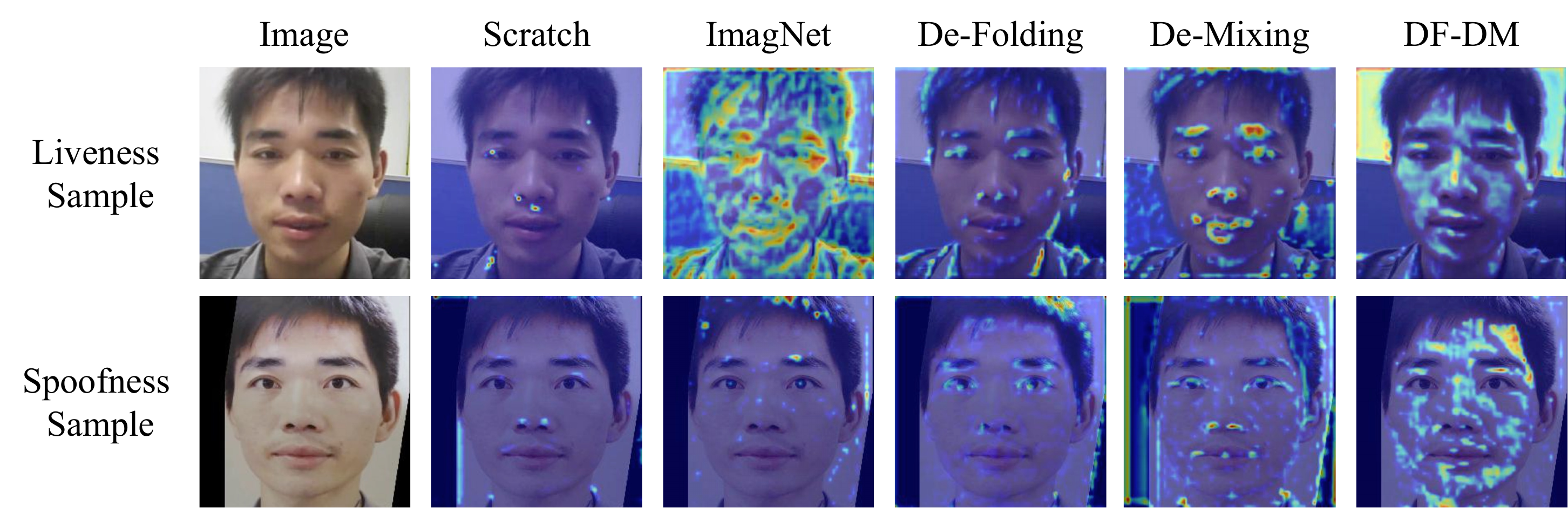}
  \caption{The Grad-CAM based visualization on CASIA-FASD  The first row shows the liveness sample and the second row presents the spoofness sample . Note that the visualized model is not trained by PAD task.
  }
  \label{fig:vis_face}
\end{figure*}
\section{Visualization Results}
To further investigate the advance of the proposed method over other baselines, we visualize the discriminative features extracted by the models with same architecture but different initialization. As shown in Fig. A.\ref{fig:vis_finger}, DF-DM can localize the adequate discriminative features for both liveness and spoofness samples. However, in the terms of De-Folding and De-Mixing, only partial features can be extracted. When compared with the pretrained model from ImageNet, DF-DM does not localize any background information as the initialized features, which indicates the effectiveness of the proposed method. When comes to the face PAD, similar performance can be observed  in Fig. A.\ref{fig:vis_face}.  Typically, for the given spoofness sample, only DF-DM can localize the effective region for PAD. 

\begin{figure*}[!htbp]
  \centering
  \includegraphics[width=.98\textwidth]{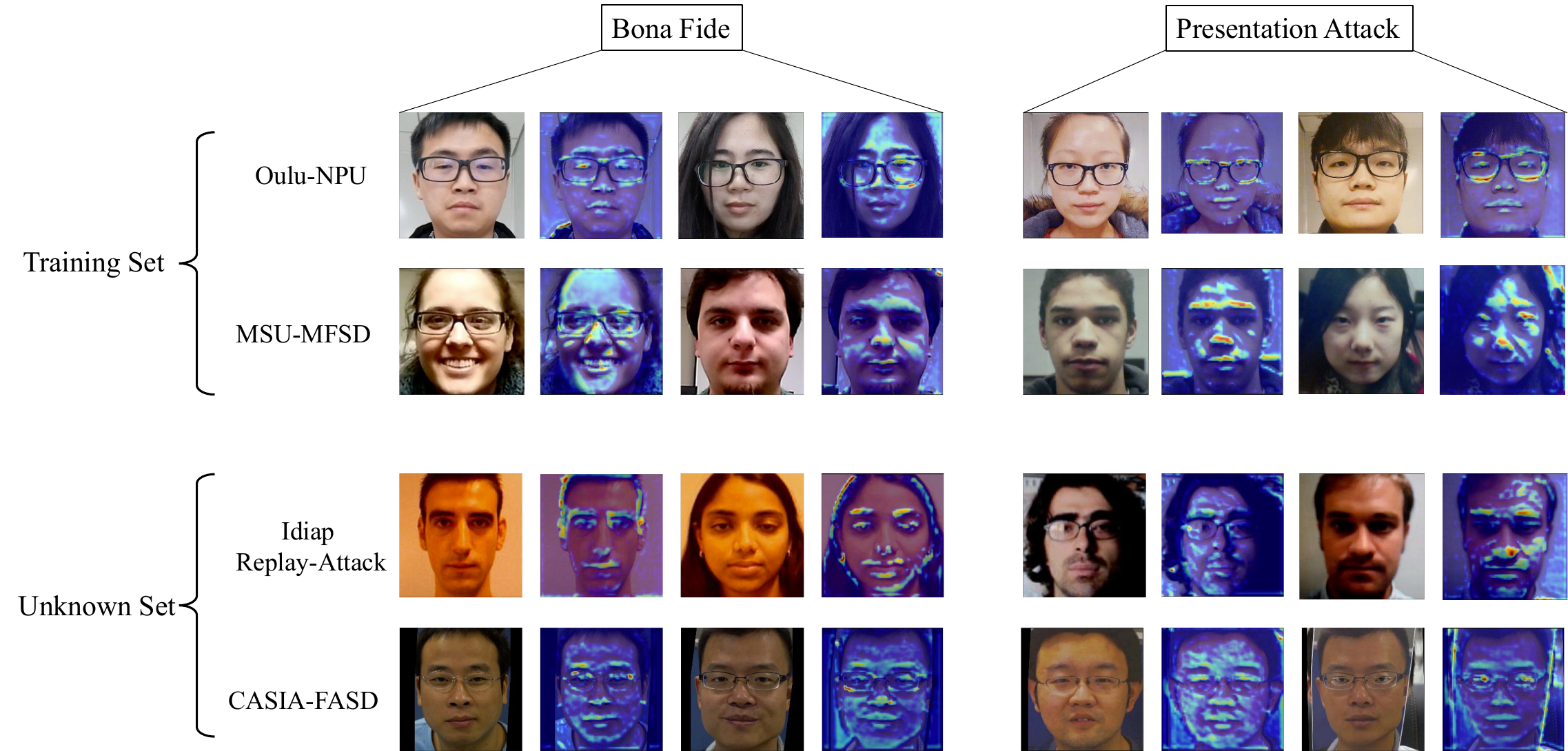}
  \caption{The Grad-CAM based visualization on Oulu-NPU, CASIA-FASD, Idiap Replay-Attack and CASIA-FASD. The model is trained by DF-DM and the training sets are Oulu-NPU and MSU-MFSD. Note that the visualized model is not trained by PAD task.
  }
  \label{fig:vis_generalization}
\end{figure*}

In order to further clarify the contribution of DF-DM for the generalization, we then visualize the discriminative features among different datasets. As shown in Fig. A.\ref{fig:vis_generalization}, although the model is trained on Oulu-NPU and MSU-MFSD, satisfactory features can also be extracted on the unknown datasets, such as Idiap Replay-Attack and CASISA-FASD. Such phenomenon indicates that DF-DM is a reasonable and ideal pre-text task for face. By only adopting limited samples for training, the model can generalize well to the other face datasets. Since the model is initialized to extract common features cross different datasets, PA detector, benefited from such initialization, can also reach to the high-level generalization.    
